\documentclass[12pt]{article}
\usepackage{math}
\usepackage{amsmath}
\usepackage{graphicx}
\usepackage{psfrag}
\usepackage{subfigure}
\usepackage[backend=biber, style=numeric]{biblatex}
\usepackage{url}
\usepackage[latin1]{inputenc} % input 8 bit ISO chars
\usepackage[T1]{fontenc} % so hyphenation works with accents
\title{\bf Multiple-Camera Tracking of Rigid Objects\thanks{This work
was sponsored by the European Commission through Esprit-IV Reactive
LTR project number 26247 VIGOR (Visually Guided Robots Using
Uncalibrated Cameras).}
}
\author{Fr\'ed\'erick Martin and Radu Horaud\\
INRIA Rh\^one-Alpes and GRAVIR-IMAG\\
655, av. de l'Europe\\ 38330 Montbonnot Saint Martin, France.
\date{}
}

\begin{document}
\maketitle
\thispagestyle{empty}
\begin{center}
International Journal of Robotics Research, 21(2), 97--113, February 2002\\
\url{https://doi.org/10.1177/027836402760475324}
\end{center}

%\tableofcontents
\begin{abstract}
In this paper we describe a method for tracking rigid objects using
one or several cameras. The tracking process consists of aligning a
3-D model representation of an object with image contours by measuring
and minimizing the image error between predicted model points and
image contours. The tracker behaves like a visual servo loop where the
internal and external camera parameters are updated at each new image
acquisition.
We study in detail the Jacobian matrix associated with this
minimization process in the presence of both point-to-point and
point-to-contour matches. We establish the minimal number of matches
that are needed as well as the singular configurations leading to a
rank-defficient Jacobian matrix. We find a mathematical link
between the point-to-point and point-to-contour cases. Based on this
link we show that the latter has the same kind of singularities than
the former. Moreover, we study multiple camera configurations which
optimize the robustness of the method in the presence of single-camera
singularities, bad, noisy, or missing data.
Extensive experiments done with a complex ship part and with up to
three cameras validate the method. In particular we show that the
tracker may well be used as a camera calibration procedure.

\end{abstract}

%\begin{keywords}

%\end{keywords}
\newpage
\section{Introduction, background, and approach}
The problem of localizing and tracking moving objects using one or
several cameras has been an active research topic. Objects fall into
several categories, from rigid, deformable, articulated and rigid, to
articulated and deformable. A common approach consists of using an
object model and of estimating the position and orientation of this
model such that some image-based error is minimized. The image error
describes the discrepancy between measured object features and
predicted model features. The ability to properly move the model such
that it optimally corresponds to the actual object depends on a number 
of factors. Roughly speaking, the 3-D parameters associated with the
object's behaviour (motion parameters, shape deformation parameters, joint
parameters for an articulated object, etc.) are related to the image
error by a Jacobian matrix. 

With only one camera there are inherent ambiguities and
singularities. Indeed, one image configuration may lead to a number of
3-D actions and Jacobian singularities may lead to no action at
all. Therefore, it may be advantageous to use several cameras instead
of one. From a practical point of view this raises the problem of
dealing with multiple camera geometry. It is more tedious to calibrate 
a multiple-camera system and to update the calibration data than to
calibrate a single camera. Moreover, the vast majority of previous
approaches consider the multi-camera system as a stereo device
requiring that each object feature is viewed in at least two images
and that image-to-image feature matches are provided. 

In this paper we propose a multi-camera based method for tracking
rigid objects. A geometric model of the tracked object is provided in
advance. The method consists of a ``visual servoing'' approach applied 
to the object model, \cite{hutchinson1996tutorial}: At each iteration of the tracking process the
image error between model and object features allows to update the
parameters associated with the position and orientation of the model.

First, we investigate the case of one camera. We establish the link
between point-to-point tracking and point-to-contour tracking. We analyse 
both the cases of fixed and varying camera internal parameters. We
study the singularities of the Jacobian matrix and we reveal the cases 
where single-camera tracking cannot be properly performed. 

Second, we investigate the case of several cameras. We show how the
single-camera case can be used to calibrate the multiple-camera
layout. We establish the tracking formulation which allows a rigid
camera layout with possibly varying internal camera parameters. This
formulation consists of running in parallel several single-camera
trackers and does not require any image-to-image correspondences. We
show how most of the single-camera singularities can be avoided by
using two or more cameras.

Third, we describe an implementation using three cameras and a complex 
rigid object. Each camera observes a different object part and
therefore the system is very robust with respect to partial occlusions 
simultaneously occurring in several images and with respect to total
occlusion of one or two cameras.

\subsection{Previous work}

The idea of using pose for tracking stems from the work of Lowe
\cite{lowe1991fitting} who used line-segment matches and the
Levenberg-Marquardt non-linear minimization method. Tracking with
variable internal camera parameters has been introduced by Kinoshita
and Deguchi \cite{kinoshita1994simultaneous} who perform visual servoing and
camera calibration {\em simultaneously}. Espiau performed an in-depths 
analysis of the convergence of visual servoing in the presence of
varying focal length \cite{espiau1993effect}. 

Harris \cite{harris1990rapid} and Armstrong and
Zisserman  \cite{armstrong1995robust} proposed to predict a model contour in the
image and to search around this prediction for image points that are
likely to lie on a matching contour. They apply this technique to the
case of straight lines and use a robust method to fit a line to the
image points. Drummond and Cipolla \cite{drummond1999b,drummond1999c,drummond2000multiple} showed
that it is possible to cast the rigid model tracking problem into a
linear problem using the Lie algebra of the rigid motion - the
kinematic screw. They suggest a series of papers dealing with a
complex 3-D object (a ship part) and with articulated objects. The
approach of Drummond and Cipolla is interesting because it relies only 
weakly on point or line matches. Instead their technique relies on
contour-to-points matching. They show that their approach can be used
for calibrating a camera. Other similar approaches to tracking using
an image prediction and searching in a window around this prediction
can be found in \cite{rasmussen2001probabilistic} and \cite{vincze2001edge}.

All these approaches to tracking consider only one camera and put
emphasis on
the data association problem. Multi-view
tracking has been barely investigated. There are several ways of using 
several cameras. One way is to consider the multi-camera setup as a
3-D sensor, capture 3-D measurements, and perform the tracking
directly in 3-D space, \cite{tonko2000model},
\cite{delamarre1999articulated}. Another way is to use the epipolar
constraint into the tracking loop itself and Lamiroy et
al. \cite{lamiroy2000b} combine the tracking equations 
with the epipolar geometry constraint. Finally, when the cameras are
far apart, there are very few image-to-image correspondences and
single camera trackers may be performed in parallel. Thompson et
al. \cite{thompson2001providing} extended the work of Harris and
of Armstrong and Zisserman to 
multiple cameras and parallel trackers. They studied the improvements
in performance using several cameras rather that using one
camera. They describe an interesting and orignial teleoperation
application where synthetic views are produced on-line thus
maintaining, at a distance, visual information about the spatial
distribution and location of polyhedral objects.

In this paper we derive an explicit algebraic expression for the
Jacobian matrix associated with point-to-contour
correspondences. Although in the past robust results were obtained
with this type of data association, none thoroughly studied the
algebraic structure of the Jacobian, the minimal data sets necessary
to run a tracker, and the possibly singular configurations which lead
to tracking failures. Based on this analysis we are able to show that
a multiple-camera approach based on point-to-contour correspondences
lead to a robust rigid-object tracker that can deal with large occlusions.

Previous work on object tracking has mainly dealt with off-line
calibrated cameras (with the notable exception of \cite{drummond2000multiple}). This means
that on-line changes (focus setting, zooming, etc.) cannot be taken
into account and they are likely to affect the accuracy of
inter-camera calibration and of the pose parameters. The possibility
to adjust these parameters during tracking is thoroughly investigated
and evaluated in this paper.

\subsection{Paper organization}
The remaining of this paper is organized as
follows. Section~\ref{section:formulation} reviews the problem of
tracking a rigid object using a perspective camera model. The
relantionship between the pose parameters (to be estimated by the
tracker) and errors associated with the internal camera parameters is
thoroughly investigated. Rigid object tracking is treated as a
minimization problem and explicit formulae are derived for two types
of image-to-model matches: point-to-point and point-to-contour
assignments. Section~\ref{section:point-point-correspondences}
analyses the singularities associated with point-to-point assignments
while section~\ref{section:point-contour-correspondences} provides a
similar analysis for point-to-contour assignments. In particular,
singular cases are revealed especially when the internal camera
parameters are allowed to vary. Multiple-camera tracking is studied in
section~\ref{section:multiple-camera} where it is shown that the
singularities associated with one camera disappear when two or more
cameras are being used. Finally section~\ref{section:experiments}
describes extensive experiments that illustrate all the variations of
the tracking method.

\section{Problem formulation}
\label{section:formulation}

\paragraph{Mathematical notations.} 
The following notations will be used throughout the paper. Bold
characters represent matrices, italic bold characters represent
vectors and roman characters represent scalars. 

Without loss of generality we consider a moving rigid object observed
by one camera. Tracking consists of a model being moved such that its apparent
position and orientation with respect to the camera corresponds to the 
real position and orientation of the object. Let
$\svect_0=(\mvect_1^0,\ldots,\mvect_k^0)$ be a set of 
model points projected onto the image based on the current prediction, 
and let $\svect$ be the
set of
corresponding image points. The relationship between the model
velocity and its apparent image velocity is:
\[
\dvv{s} = \Jmat \Tvect
\]
where $\Tvect$ is the kinematic screw associated with the 3D
model motion. The error between the predicted model position and the
actual object position may be measured from their image projections:
\[
\evect = \Cmat \left( \svect_0 - \svect \right)
\]
Where \( \Cmat \) is combination matrix allowing to consider more
measurements than the number of degrees of freedom (six in this case
-- three for the rotational velocity vector and three for the
translational velocity vector). A common choice which insures
convergence is to set $\Cmat=\Jmat\tp$.

The objective is to move the model such that the image error decreases:

\[
\lambda \Jmat\tp (\svect_0 - \svect) = - \Jmat\tp \dvv{s} = -  \Jmat\tp\Jmat \Tvect
\]
which allows as solution:
\begin{equation}
	\Tvect =  \lambda \left(\Jmat\tp\Jmat\right)\inverse\Jmat\tp (\svect - \svect_0)
	\label{eq:visual-servoing}
\end{equation}

This is the basic visual
servoing equation which is solved in order to maintain
$\svect-\svect^\star$ as small as possible. In order to apply this
formulation to the tracking problem, one may approximate the
velocity screw by:
\[
\Tvect = \frac{d\xvect}{dt} = \frac{\xvect - \xvect_0}{t-t_0}
\]
Hence, the equation above allows the incremental update of the pose
parameters $\xvect$:
\begin{equation}
\xvect = \xvect_0 + \lambda (t-t_0) \left(\Jmat\tp\Jmat\right)\inverse\Jmat\tp (\svect
- \svect_0)
\label{eq:tracking-pose}
\end{equation}

\subsection{Camera model}

The model-based object tracking just described assumes a pinhole camera model,
i.e., an object point projects onto an image point using the well known 
camera projection matrix:
\begin{equation}
\tilde{\mvect} = \Kmat \left[ \begin{array}{cc} \Rmat & \tvect
\end{array} \right] \tilde{\Mvect} 
\label{eq:camera-projection}
\end{equation}

where $\tilde{\mvect}$ is a homogeneous 3-vector describing the
projective
coordinates of 
an image point, $\Kmat$ is the matrix of internal camera parameters,
$\Rmat$ is a rotation matrix, $\tvect$ is a translation vector, and
$\tilde{\Mvect}$ is a homogeneous 4-vector describing the coordinates of a
model point.

The internal camera parameters are $f$ (horizontal and vertical scale
factor, or focal length) and $u_0$ and $v_0$ (image coordinates of
optical center):
\[
\Kmat = \left[ \begin{array}{ccc}
f & 0 & u_0 \\
0 & f & v_0 \\
0 & 0 & 1
\end{array} \right]
\]
This is a simplified camera model since it is well known that the
horizontal to vertical scale factor can be estimated off-line, once
for ever, and the value thus estimated is stable over time and over
changes in camera settings.

\subsection{The image Jacobian}

The 2-vector $\mvect$ describes the pixel coordinates associated with
the projection $\tilde{\mvect}$. By taking the time derivatives we
obtain the classical relationship:

\begin{equation}
\frac{d\mvect}{dt} = \Jmat_m \left( \begin{array}{c}
v_x \\ v_y \\ v_z \\ w_x \\ w_y \\ w_z \\  \frac{df}{dt} \\
\frac{du_0}{dt} \\ \frac{dv_0}{dt} \end{array} \right)
\label{eq:im-Jac-1}
\end{equation}
with the kinematic screw $\Tvect\tp=(v_x \; v_y \;v_z \;w_x \;w_y \;w_z)$ and with:
\begin{equation}
\Jmat_m = f
\left[ 
\begin{array}{rrccccccc}
-\frac{1}{z} & 0 & \frac{x}{z^2} & \frac{xy}{z^2} &
-\left(1+\frac{x^2}{z^2}\right) & \frac{y}{z} & \frac{x}{zf} & \frac{1}{f}  & 0  \\
0 & -\frac{1}{z} & \frac{y}{z^2} & \left(1+\frac{y^2}{z^2}\right) &
- \frac{xy}{z^2}
 & \frac{-x}{z} &  \frac{y}{zf} & 0 &\frac{1}{f} 
\end{array} \right]
\label{eq:im-Jac-m}
\end{equation}

where $(x\; y\; z)$ are the Euclidean coordinates of point $M$ expressed
in camera frame.

\subsection{Camera model errors}

The camera parameters may be unknown, partially known, or badly known. One
important feature of the tracking algorithm described below is that it 
can accommodate with rough estimates of these parameters which are
updated during tracking. 

We analyse the effect of badly known camera parameters onto the pose
parameters. Let $\Kmat$ be the exact camera parameters and let $\Rmat$ 
and $\tvect$ be the pose parameters associated with $\Kmat$ and with
the point correspondences $\mvect\leftrightarrow\Mvect$. Let $\hatmat{K}$
be an estimation of the true camera parameters. With these estimated
parameters one can associate estimated pose parameters $\hatmat{R}$ and
$\hatvec{t}$. The projection equation holds in both these cases:
\[
\tilde{\mvect} = \hatmat{K}
\left[ \hatmat{R} \; \hatvec{t} \right]
\tilde{\Mvect} = \Kmat 
\left[ \begin{array}{cc} \Rmat & \tvect \end{array} \right]
\tilde{\Mvect}
\]
We have:
\[
\tilde{\mvect} = \Kmat
\left[ \begin{array}{cc} \Kmat\inverse\hatmat{K}\hatmat{R} &
\Kmat\inverse\hatmat{K}\hatvec{t} \end{array} \right]
\tilde{\Mvect}
\]
By writing $\hatmat{K}=\Kmat+d\Kmat$ and by first-order Taylor expansion
we obtain, \cite{horaud2000stereo}
\begin{equation}
\Kmat{'}\inverse\Kmat = \Imat - \left[ \begin{array}{ccc}
\frac{df}{f} & 0 & \frac{du_0}{f} \\
0 & \frac{df}{f} & \frac{dv_0}{f} \\
0 & 0 & 0
\end{array} \right] = \Imat + \Kmat_\varepsilon
\label{eq:focale}
\end{equation}
The pose parameters are affected by these internal parameter errors as 
follows:
\begin{eqnarray*}
\hatmat{R}& =& \Rmat - \Kmat_\varepsilon\Rmat\\
\hatvec{t}& =& \tvect - \Kmat_\varepsilon\tvect
\end{eqnarray*}
The estimation of $f$, $u_0$ and $v_0$ is affected by the value of
$f$. 
For small focal
lengths the internal parameters have equal importance. For large focal 
lengths, the accuracy of $f$ is intrinsically more crucial than the
accuracy of the optical center. If the focal center lies at
approximatively the image
center, the value of $f$ is, on an average, 4-5 times greater than the 
values of $u_0$ and $v_0$. Therefore, the ability to track in the
presence of variable focal length, estimate and correct its value on
line is an important feature.

\subsection{Tracking as a minimization problem}

We consider now a more general tracking problem where the object is
observed simultaneously by several cameras with possibly varying
internal camera parameters. Let $\qvect$ denote the state-vector
associated with such a setup: this vector encapsulates the pose
parameters of the object with respect to a global camera centered
frame as well as the internal parameters associated with the camera
models. Therefore, the model predictions $\svect_0$ are a function of
$\qvect$.

One must distinguish between two cases:
\begin{enumerate}
\item Tracking based on point-to-point correspondences
\item Tracking based on point-to-contour correspondences
\end{enumerate}

\begin{figure}
%\centerline{\includegraphics[width=0.8\columnwidth]{p2c-matches.pdf}}
\centerline{\resizebox{0.6\textwidth}{!}{  \includegraphics{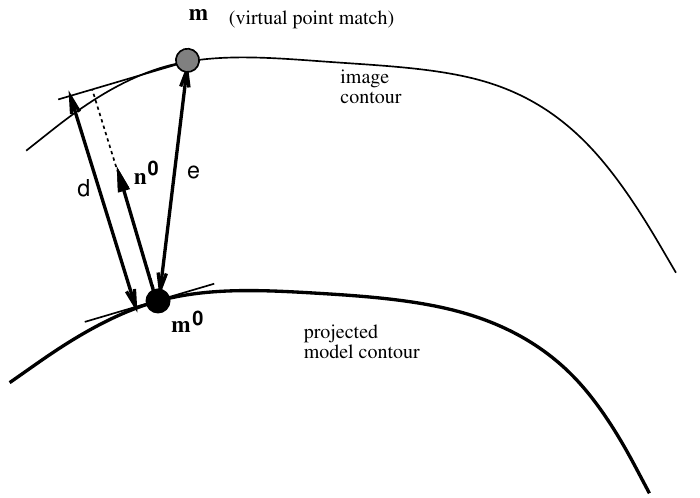}  }}
\caption{\label{fig:matches} A model contour is predicted in the image and a point
location $\mvect^0$ along this contour is considered. For point-to-point 
correspondences, it is sufficient to search for a matching point
$\mvect$ end to determine the distance $e$ between these
points. For point-to-contour correspondences the strategy consists of
raising a normal to the model contour at location $\mvect^0$, finding
the intersection of the line supporting the normal vector with the
matching image contour, and determining the distance from $\mvect^0$ to
the image contour. The algebraic expression of this true
point-to-contour distance can be approximated by
$d=\nvect^0\cdot(\mvect-\mvect^0)$.}
\end{figure}

These two cases are illustrated on Figure~\ref{fig:matches}. In the
first case (point-to-point) the function to be minimized may be
written as the sum of the Euclidean distances from the predicted model 
point $\mvect_i^0$ to its image match $\mvect_i$:
\[
Q_1(\qvect) = \frac{1}{2} \sum_{i=1}^{n}\|
\mvect_i^0(\qvect) -\mvect_i \|^2
= \frac{1}{2} \| \svect - \svect_0 \|^2
\]

Therefore, the general form of the minimization problem is, in this
case:
\begin{equation}
Q_1(\qvect) = \frac{1}{2} \rvect(\qvect)\tp \rvect(\qvect)
\label{eq:minimization-ptop}
\end{equation}
with:
\[ \rvect = \svect - \svect_0 \]

In practice it is well known that it is difficult to obtain
point-to-point correspondences. A current approach is to project a
model contour onto the image using the current pose and camera
parameters, search in a direction
normal to the predicted model contour, find a corresponding image
point lying onto a contour, and measure the distance to
this point \cite{marchand1999robust},
\cite{drummond1999a}. For a projected model point $\mvect_i^0$ and a normal
direction $\nvect_i^0$, this distance is approximated by, e.g.,
Figure~\ref{fig:matches}:
\[
d_i (\qvect) = \nvect_i^0(\qvect) \cdot (\mvect_i^0 (\qvect) - \mvect_i)
\]

Notice that this formula is exact for straight-line contours and hold
only approximatively for curved contours. 
The function to be minimized may be
written as:

\begin{equation}
Q_2(\qvect) = \frac{1}{2}\left(\Nmat(\qvect)\rvect(\qvect)\right)\tp
\Nmat(\qvect)\rvect(\qvect) = \frac{1}{2} \sum_{i=1}^{n} d_i^2
\label{eq:minimization-ptoc}
\end{equation}

Matrix $\Nmat$ of size $n\times 2n$ contains the normals to the
projected 

model
contours. Its rank is equal to $n$:
\begin{equation}
\Nmat = \left[
	\begin{array}{ccccccc}
	n_{1u} & n_{1v} & 0 & 0 & \ldots & 0 & 0 \\
 	0 & 0 & n_{2u} & n_{2v} & \ldots & 0 & 0 \\
	\vdots &&&&&& \\
	0 & 0 & 0 & 0 & \ldots & n_{nu} & n_{nv}
	\end{array}
	\right]
\label{eq:N-definition}
\end{equation}

The first- (Jacobian) and second-order (Hessian) derivatives of $Q_1$ with
respect to $\qvect$  
are:
\[
\frac{dQ_1}{d\qvect} = \Jmat_1\tp(\qvect) \rvect(\qvect)
\]
\[
\frac{d^2Q_1}{d\qvect^2} = \Jmat_1\tp(\qvect) \Jmat_1(\qvect) +
\frac{d^2\rvect}{d\qvect^2}\rvect(\qvect)
\]

Where the Jacobian is defined by:
\[
\Jmat_1 = \frac{d\rvect(\qvect)}{d\qvect}
\]
We derive now the Jacobian matrix for
point-to-contour correspondences:
\begin{eqnarray*}
\Jmat_2 &=& \frac{d\left(\Nmat(\qvect)\rvect(\qvect)\right)}{d\qvect} \\
        &=& \Nmat(\qvect) \Jmat_1 +
\left(\sum_{i=1}^{i=k}\frac{\partial\Nmat(\qvect)}{\partial q_i}
\right) \rvect(\qvect)
\end{eqnarray*}

Matrix $\Nmat$ contains the normals to the contour. Its derivatives
may therefore be considered as second-order terms and therefore they
can be neglected in the expression of the Jacobian. The latter becomes:

\begin{equation}
\Jmat_2 = \Nmat(\qvect) \Jmat_1 (\qvect)
\label{eq:Jacobian-point-to-contour}
\end{equation}

The rank of $\Jmat_2$ is related to the rank of $\Jmat_1$ by the
formula:
\begin{equation}
r(\Jmat_2) \leq \min(\rank (\Jmat_1), \rank (\Nmat) ) = \min(\rank (\Jmat_1),n)
\label{eq:rank-theorem-1}
\end{equation}
because the rank of $\Nmat$ is equal to $n$. This will be useful for
analysing the convergence of tracking.

A common way to solve this problem is to consider the Gauss-Newton
approximation of the Hessian, $\frac{d^2Q}{d\qvect^2} =
\Jmat\tp(\qvect) \Jmat(\qvect)$ and to perform a number of quadratic
iterations. One such quadratic iteration writes for $\Jmat_1$ (where
there is a similar expression for $\Jmat_2$):
\begin{equation}
\qvect^\diamond = \qvect - \left(\Jmat_1\tp(\qvect)
\Jmat_1(\qvect)\right)\inverse \Jmat_1\tp(\qvect)  \rvect (\qvect)
\label{eq:general-tracking}
\end{equation}

In practice and due to time constraints, only one iteration is
performed each time a new image is grabbed. Therefore, there are two
situations:
\begin{itemize}
\item The object is static with respect to the camera in which case
the method behaves like a non-linear pose estimation technique or
\item The object moves in which case the method acts like an object
tracker maintaining a pose estimation as close as possible to a
previously computed solution.
\end{itemize}

Notice that the Gauss-Newton iteration given by
eq.~(\ref{eq:general-tracking}) is strictly equivalent to
eq.~(\ref{eq:tracking-pose}). Both these equations require the 
Jacobian matrix to be of rank equal to the dimension of the state vector
$\qvect$. In the sequel we analyse cases for which the Jacobian matrix 
has singularities. An alternative way to avoid such singularities is to consider
trust-region minimization methods for solving
eq.~(\ref{eq:minimization-ptop}) or eq.~(\ref{eq:minimization-ptoc}),
\cite{lowe1991fitting,lowe1992robust}, 
\cite{phong1995object}.

\section{Point-to-point correspondences}
\label{section:point-point-correspondences}

The relationship between image measurements and incremental rigid
motions allowing the model-to-object alignment is encapsulated in the
Jacobian matrices $\Jmat_1$ and $\Jmat_2$ mentioned above. We
distinguish the following cases: 
\begin{enumerate}
\item Fixed internal camera parameters
\item Varying focal length
\item Varying internal camera parameters (focal length and optical center)
\end{enumerate}

In the case of point-to-point correspondences, the Jacobian is of
dimensions $2n\times k$ where $n$ is the number of point
correspondences and $k$ is the number of parameters to be recovered (the 
dimension of $\qvect$). For the general case (varying internal
parameters) and for one point-to-point match it has the algebraic
structure given by eq.~(\ref{eq:im-Jac-m}). 

In the case of point-to-contour correspondences, the Jacobian is of
dimension $n\times k$ where $n$ is the number of point-to-contour
correspondences.

In order to study potential failures of the tracking method and,
possibly, avoid them we must analyse the 
rank of the Jacobian for the minimum number of points and contours, and 
determine geometric configurations for which the Jacobian is rank deficient.

\subsection{Fixed internal camera parameters}

There are six parameters 
to be estimated in this case (three for rotation and three for
translation) and hence the least number of correspondences is equal to 
3. This case was thoroughly studied within the convergence analysis of 
visual servoing methods. The Jacobian is rank defficient when (i)~the
image points are collinear or when (ii)~the three model points together
with the optical center lie on a cylinder:
\begin{equation}
\Jmat_1 = \left[ \begin{array}{rrcllc}
\vdots &&&&&\\
-\frac{1}{z_i} & 0 & \frac{x_i}{z_i^2} & \frac{x_iy_i}{z_i^2} &
-\left(1+\frac{x_i^2}{z_i^2}\right) & \frac{y_i}{z_i}  \\
0 & -\frac{1}{z_i} & \frac{y_i}{z_i^2} & \left(1+\frac{y_i^2}{z_i^2}\right) &
- \frac{x_iy_i}{z_i^2}
 & \frac{-x_i}{z_i} \\
\vdots &&&&&
\end{array} \right]
\label{eq:Jacobian-ptp-fixed}
\end{equation}

\subsection{Varying focal length} 
\label{subsection:varying-focal-1}

In this case there are seven parameters
to be estimated. Therefore, a minimum of four correspondences are
required. Due to the structure of the Jacobian matrix, a planar 
configuration with four (or more) points lying in a plane parallel to
the image plane 
gives rise to a Jacobian of rank equal to 6. Indeed, for such a
configuration, the depth parameters $z_i$ are the same for all the
points and the third column is proportional to the last column in the
Jacobian matrix:
\begin{equation}
\Jmat_1 = f \left[ \begin{array}{rrcllccc}
\vdots &&&&&&\\
-\frac{1}{z_i} & 0 & \frac{x_i}{z_i^2} & \frac{x_iy_i}{z_i^2} &
-\left(1+\frac{x_i^2}{z_i^2}\right) & \frac{y_i}{z_i} & \frac{x_i}{z_if}  \\
0 & -\frac{1}{z_i} & \frac{y_i}{z_i^2} & \left(1+\frac{y_i^2}{z_i^2}\right) &
- \frac{x_iy_i}{z_i^2}
 & \frac{-x_i}{z_i} & \frac{y_i}{z_if} \\
\vdots &&&&&&
\end{array} \right]
\label{eq:Jacobian-ptp-focal}
\end{equation}

This means that it is not 
possible to simultenously track a planar object and estimated the
focal length when this 
object lies in a plane parallel (or almost parallel) to the image
plane. This singularity extends to the case where an arbitrary 3-D object is at a
distance from the camera such the weak-perspective approximation
becomes valid. In fact this configuration is the only known
situation which does not allow the distinguish between variations in
depth and variations in focal length.

\subsection{Varying internal camera parameters} 

In this case there are 
9 parameters to be estimated (provided that the camera's
horizontal-to-vertical scale ratio is known) and hence the Jacobian
must be of rank at least equal to 9. Again, a fronto-parallel planar
object gives rise to a rank-defficient Jacobian matrix for the same algebraic
reasons as above.
\begin{equation}
\Jmat_1 = f \left[ \begin{array}{rrcllccccc}
\vdots &&&&&&&&\\
-\frac{1}{z_i} & 0 & \frac{x_i}{z_i^2} & \frac{x_iy_i}{z_i^2} &
-\left(1+\frac{x_i^2}{z_i^2}\right) & \frac{y_i}{z_i} & \frac{x_i}{z_if} 
 &
\frac{1}{f}  & 0  \\
0 & -\frac{1}{z_i} & \frac{y_i}{z_i^2} & \left(1+\frac{y_i^2}{z_i^2}\right) &
- \frac{x_iy_i}{z_i^2}
 & \frac{-x_i}{z_i}  & \frac{y_i}{z_if} &
0 & \frac{1}{f}  \\
\vdots &&&&&&&&
\end{array} \right]
\label{eq:Jacobian-ptp-all}
\end{equation}

\section{Point-to-contour correspondences}
\label{section:point-contour-correspondences}

The relationship between $\Jmat_2$ and $\Jmat_1$ established in
the previous section -- $\Jmat_2=\Nmat\Jmat_1$ -- allows analysis
of the singularities associated  
with this type of correspondences. Unlike the point-to-point case, each
point-to-contour match contributes with one row in the Jacobian
matrix $\Jmat_2$. Therefore, twice more point-to-contour
than point-to-point correspondences are necessary. At
first glance this may appear as a disadvantage. In practice, it is
easier to establish point-to-contour correspondences and hence a
method based on such correspondences is likely to be more robust. It
is often the case that several points match a single contour. It will
be shown that in the case of linear contours, the minimum number of
image lines required by the point-to-contour tracker is equal to the number
of image points required by the point-to-point tracker.

In order to implement a tracker based on point-to-contour
correspondences, the rank of the Jacobian matrix $\Jmat_2$ must be
equal to 9 (position, orientation, and varying internal camera
parameters), 7 (position, orientation, and varying focal length), or
to 6 (position and orientation). Consequently a minumum of 9, 7, or 6 matches are
necessary. We analyse these situations in reverse order.

\subsection{Fixed internal camera parameters}

\begin{figure}
\centerline{\resizebox{0.5\textwidth}{!}{  \includegraphics{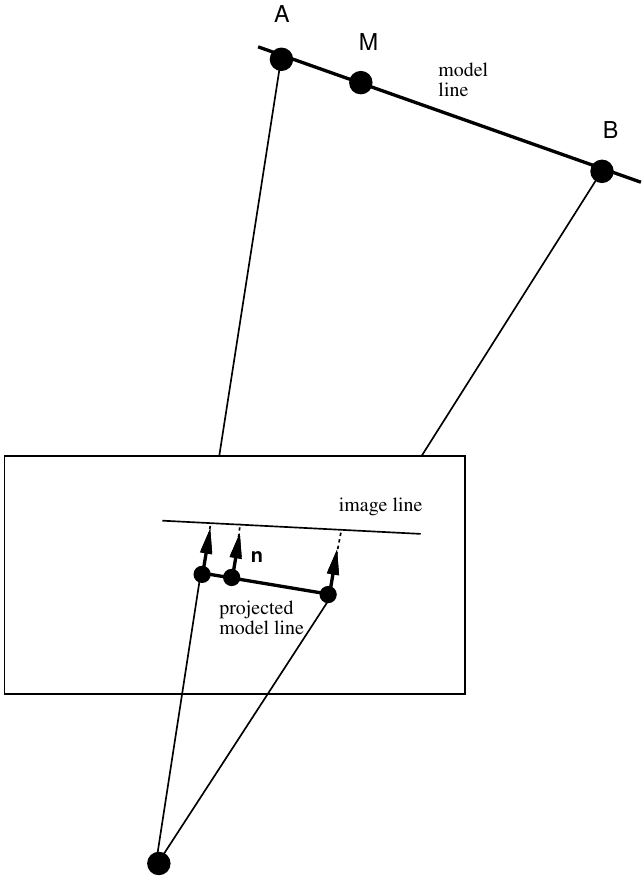}  }}
\caption{\label{fig:line-projection} This figure shows a match between a model point (lying onto a 
straight-line and an image line, i.e., a point-to-line correspondence.}
\end{figure}

Without loss of generality we consider a set of colinear model points, 
i.e., model points lying on linear edges. Let $A$ and $B$ be
two points belonging to a linear edge with camera coordinates
$\Avect=(x_A \; y_A \;
z_A)\tp$ and $\Bvect=(x_B \; y_B \; z_B)\tp$. A third point $M$ has as
coordinates
$\Mvect = \Avect + \lambda (\Bvect - \Avect)$
or:
\begin{eqnarray*}
x_M &=& x_A + \lambda (x_B - x_A) \\
y_M &=& y_A + \lambda (y_B - y_A) \\
z_M &=& z_A + \lambda (z_B - z_A) 
\end{eqnarray*}
where $\lambda$ is a free parameter, e.g.,
Figure~\ref{fig:line-projection}.

These three model points project onto the image producing an image
line and let $\nvect=(n_u \; n_v)\tp$ be the  normal 
to this line. The pixel coordinates of $\nvect$ can be computed from
the coordinates of vectors $\Avect$ and $\Bvect$ \cite{phong1995object}:
\[
\nvect = \frac{\pvect}{ \|\pvect\|} \; \mbox{with} \; \pvect = 
\left( \begin{array}{c} p_u \\ p_v \end{array} \right) =
\left(
\begin{array}{c} y_B z_A - y_A z_B \\ x_A z_B - x_B z_A \end{array} \right)
\]

Matrix $\Nmat$, containing the image normals to the predicted model
contour, has a simple expression for three collinear points:
\[
\Nmat = \frac{1}{ \|\pvect\|} \left[ \begin{array}{cccccc}
p_u & p_v & 0 & 0 & 0 & 0 \\
0 & 0 & p_u & p_v & 0 & 0 \\
0 & 0 & 0 & 0 & p_u & p_v 
\end{array} \right]
\]

The Jacobian associated with this configuration -- three
colinear points matching an image contour -- writes, i.e.,
eqs.~(\ref{eq:N-definition}), (\ref{eq:Jacobian-ptp-fixed}), and
(\ref{eq:Jacobian-point-to-contour}): 

\begin{eqnarray}
\label{eq:Jacobian-3p-1line}
& \Jmat_2 = \frac{\pvect}{ \|\pvect\|}
\left[  \begin{array}{ccc}
\frac{1}{z_A^2} & 0 & 0 \\
0 & \frac{1}{z_B^2} & 0 \\
0 & 0 & \frac{1}{z_M^2}
\end{array} \right]
\left[  \begin{array}{lll}
-p_uz_A & -p_vz_A & p_ux_A+p_vy_A  \\
-p_uz_B & -p_vz_B & p_ux_B+p_vy_B  \\
-p_uz_M & -p_vz_M & p_ux_M+p_vy_M 
\end{array} \right. &  \\
& \left. \begin{array}{lll}
p_ux_Ay_A+p_v(z_A^2+y_A^2) & -p_u(z_A^2 + x_A^2) - p_vx_Ay_A & p_uy_Az_A - p_vx_Az_A \\
p_ux_By_B+p_v(z_B^2+y_B^2) & -p_u(z_B^2 + x_B^2) - p_vx_By_B & p_uy_Bz_B - p_vx_Bz_B \\
p_ux_My_M+p_v(z_M^2+y_M^2) & -p_u(z_M^2 + x_M^2) - p_vx_My_M & p_uy_Mz_M - p_vx_Mz_M 
\end{array}  \right] & \nonumber
\end{eqnarray}

By substituting the coordinates of $\Mvect$ and $\pvect$ with their
expressions as functions of the coordinates of $\Avect$ and $\Bvect$
we obtain that the third row of the 3$\times$6 matrix above is a
linear combination of the first and second rows. Therefore the rank of 
$\Jmat_2$ is equal to 2. In other terms, points $M_i$ which are collinear 
with $A$ and $B$ are redundant. Regardless of the number of
points the rank of the Jacobian associated with $n$ collinear points
matching an image line is at most equal to 2.

\subsection{Varying focal length}
\label{section:varying-focal-2}

The Jacobian is a 3$\times$7 matrix in this case. The first six 
columns are identical to eq.~(\ref{eq:Jacobian-3p-1line}). The last
column is obtained by combining the expression of matrix $\Nmat$ with
eq.~(\ref{eq:Jacobian-ptp-focal}) and is equal to:
\[
\begin{array}{c}
\frac{z_A}{f}(p_ux_A + p_vy_A) \\
\frac{z_B}{f}(p_ux_B + p_vy_B) \\
\frac{z_M}{f}(p_ux_M + p_vy_M) 
\end{array}
\]

As before, by substitution, we obtain the third row as a linear
combination of the first two. Therefore the rank of the 3$\times$7
Jacobian matrix is equal to 2.

\subsection{Varying internal camera parameters}

The Jacobian is a 3$\times$9 matrix in this case. The first seven
columns are identical to the previous case. The last two columns are
obtained by combining the expression of matrix $\Nmat$ with
eq.~(\ref{eq:Jacobian-ptp-all}) and yield the following 3$\times$2
block:

\[
\begin{array}{cc}
\frac{z_A^2p_u}{f} & \frac{z_A^2p_v}{f} \\
\frac{z_B^2p_u}{f} & \frac{z_B^2p_v}{f} \\
\frac{z_M^2p_u}{f} & \frac{z_M^2p_v}{f} 
\end{array}
\]

Again, the rank of this matrix is equal to 2. 

\section{Multiple camera tracking}
\label{section:multiple-camera}

Single-camera tracking suffers from three basic limitations: 
\begin{itemize}
\item As explained above, there are
cases where the Jacobian matrix is rank-defficient and hence tracking
cannot be performed (planar objects lying in a fronto-parallel plane);
\item Due to occlusions, bad imaging conditions, and so forth, there
may not be enough features available in one image, and 
\item While the object moves it may, totally or partially, disappear
from the camera field of view.
\end{itemize}
	
There are two options enabling one to introduce multiple-camera
tracking. The first possibility is to make use of multiple-camera geometry,
as described in several books \cite{faugeras1993three},
\cite{hartley2000multiple}, \cite{luong2001geometry}, and which implies
that all cameras have a large field of view in common. The second
possibility is to execute single-camera trackers in parallel while
making explicit the object's rigidity. 

Indeed, if each one of the cameras looks at a
different object part, the whole object is nevertheless rigid and
geometric knowledge allows individual calibration of each one of the
cameras. Once this calibration step has been performed, all the
cameras track the object in parallel estimating its motion in a common 
reference frame. Such a tracker allows, one feature to be seen by one
camera, another feature to be seen by another camera, etc. 

Since the latter method does not use any form of stereo correspondence, the
cameras can be quite far apart thus insuring that a planar piece of
the object
appearing in one's camera fronto-parallel plane is viewed slanted
by the other cameras. Moreover, even if each camera sees planar
features in its own fronto-parallel plane, the Jacobian matrix resulting from all 
cameras is not rank defficient.

\subsection{The multiple camera Jacobian matrix}

Without loss of generality, we consider that each camera has been
individually calibrated using the tracker described in the previous
section in conjunction with a static object. This means that both
internal and external (pose) parameters are available for each
camera and that the position and orientation of each camera is known
in a common reference frame.

Therefore the multiple camera tracker is left with six parameters to be
estimated: the position and orientation of the object with respect to
the common reference frame.

With the same notations as above, we generalize the
point-to-point Jacobian matrix such that the
reference frame is an arbitrary 
frame rather than the camera frame. For camera $k$ the Jacobian
writes:

\begin{equation}
	\Jmat_{1}^{k} = 
 \left[ \begin{array}{ccccccc}
\vdots &&&&\\
-\frac{1}{z_{ik}} & 0 & \frac{x_{ik}}{z_{ik}^2} & \frac{x_{ik}y_{ik}}{z_{ik}^2} &-\left(1+\frac{x_{ik}^2}{z_{ik}^2}\right) & \frac{y_{ik}}{z_{ik}} \\
0 & -\frac{1}{z_{ik}} & \frac{y_{ik}}{z_{ik}^2} & \left(1+\frac{y_{ik}^2}{z_{ik}^2}\right) &
- \frac{x_{ik}y_{ik}}{z_{ik}^2}
 & \frac{-x_{ik}}{z_{ik}} \\
\vdots &&&&
\end{array} \right]
\left[\begin{array}{cc} \Rmat_k & [\tvect_k]_{\times}\Rmat_k \\ 0 &
\Rmat_k \end{array} \right] \\ 
\label{eq:multiple-Jacobian}
\end{equation}
where $\Rmat_k$ and $\tvect_k$ are the rotation and translation
mapping 3-D coordinates of model points from a global common reference frame
to the camera frame. The notation $[]_\times$ stands for the skew
symmetric matrix associated with a 3-vector, $x_{ik}$, $y_{ik}$, and
$z_{ik}$ represent the coordinates of a model point $i$ in camera
frame $k$. 

The multiple camera Jacobian matrix writes:

\begin{equation}
\Jmat_1 = \left[ \begin{array}{c}
\Jmat_1^1 \\ \vdots \\ \Jmat_1^k \\ \vdots 
\end{array} \right]
\label{eq:multiple-Jacobian-1}
\end{equation}

There is a similar expression for the point-to-contour multiple-camera 
Jacobian matrix:

\begin{equation}
\Jmat_2 = \left[ \begin{array}{c}
\Nmat^1 \Jmat_1^1 \\ \vdots \\ \Nmat^k \Jmat_1^k \\ \vdots 
\end{array} \right]
\label{eq:multiple-Jacobian-2}
\end{equation}

To conclude, with two or more cameras the tracker is well
conditioned. Such configurations as cameras with parallel image planes 
may cause problems ($\Rmat_k=\Imat$ in
eq.~(\ref{eq:multiple-Jacobian})) but such configurations can easily be
avoided. Moreover,
they are not of very useful in practice. 

\section{Experiments}
\label{section:experiments}

The single and multiple camera tracking methods described above were
used to locate complex ship parts for subsequent ship building operations 
such as welding \cite{lamiroy2000a}.

Throughout the experiments we used two RGB 
frame grabbers in conjuction with 6 black-and-white cameras. 
This hardware setup guarantees image grabbing synchronization.
The
cameras are thus interfaced to a PC/Pentium3
(733MHz) running
under Linux. The image size is 768$\times$574 pixels. The tracker runs
at 25Hz with one camera, at 17Hz with two cameras, at 12Hz with
three cameras, ... and at 6Hz with six cameras. This corresponds to
the object model described below which is relatively complex. 

We found that the most 
practical configurations use 2 to 4 cameras. These computation speeds
correspond to one iteration of a tracker per image. Performing several 
iterations per image has theoretical advantages in the sense that
convergence of the tracker is guaranteed. However, it is undesirable
in practice because it degrades the speed and therefore it limits the
system to slow moving objects.

The object to be
located is modeled with both straigth and curvilinear edges. There are
approximatively 300 model edges. Both these
edges (straight and curvilinear) are approximated by piecewise polygonal lines formed of short
segments: Each edge is sampled such that, on an
average there are 10 points per model edge.

\subsection{Finding point-to-contour correspondences}

One of the most crucial steps of any tracking method is to establish
matches (or correspondences). In our case we must associate model
edges with image contour points. Such associations are easier to find than
the classical point-to-point, line-to-line, or contour-to-contour
matches. The method described below is noise-, error-,  and
occlusion-tolerant because it eliminates any model-to-image data
associations that are likely to lead to an ambiguous solution.

\begin{figure}
%\centerline{\includegraphics{Hidden.pdf}}
\centerline{\resizebox{0.6\textwidth}{!}{  \framebox{\includegraphics{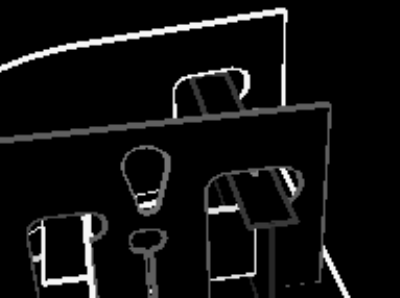}}  }}
\caption{This figure shows a z-buffer associated with an image from a
predicted point of view. Black indicates that there is no visible
edge. Each visible edge is indexed with a different colour.}
\label{fig:zbuffer}
\end{figure}

The first step in the model-to-image data association method is to
predict model edges that are visible in the image. Using the current
pose of the model with respect to one camera, a z-buffer is produced
from the object model (represented
under the VRML 2.0 description format). Z-buffering has linear
complexity in the number of model primitives. Each visible model-edge
has a color index associated with it while the hidden edges together
with the model faces are coloured in black. Hence it is
straigthforward to determine weather a point belonging to a model edge 
is visible or not: It is sufficient to index its color. This model
representation is illustrated on Figure~\ref{fig:zbuffer}.

\begin{figure}
%\centerline{\includegraphics{search-contour.pdf}}
\centerline{\resizebox{0.6\textwidth}{!}{  \includegraphics{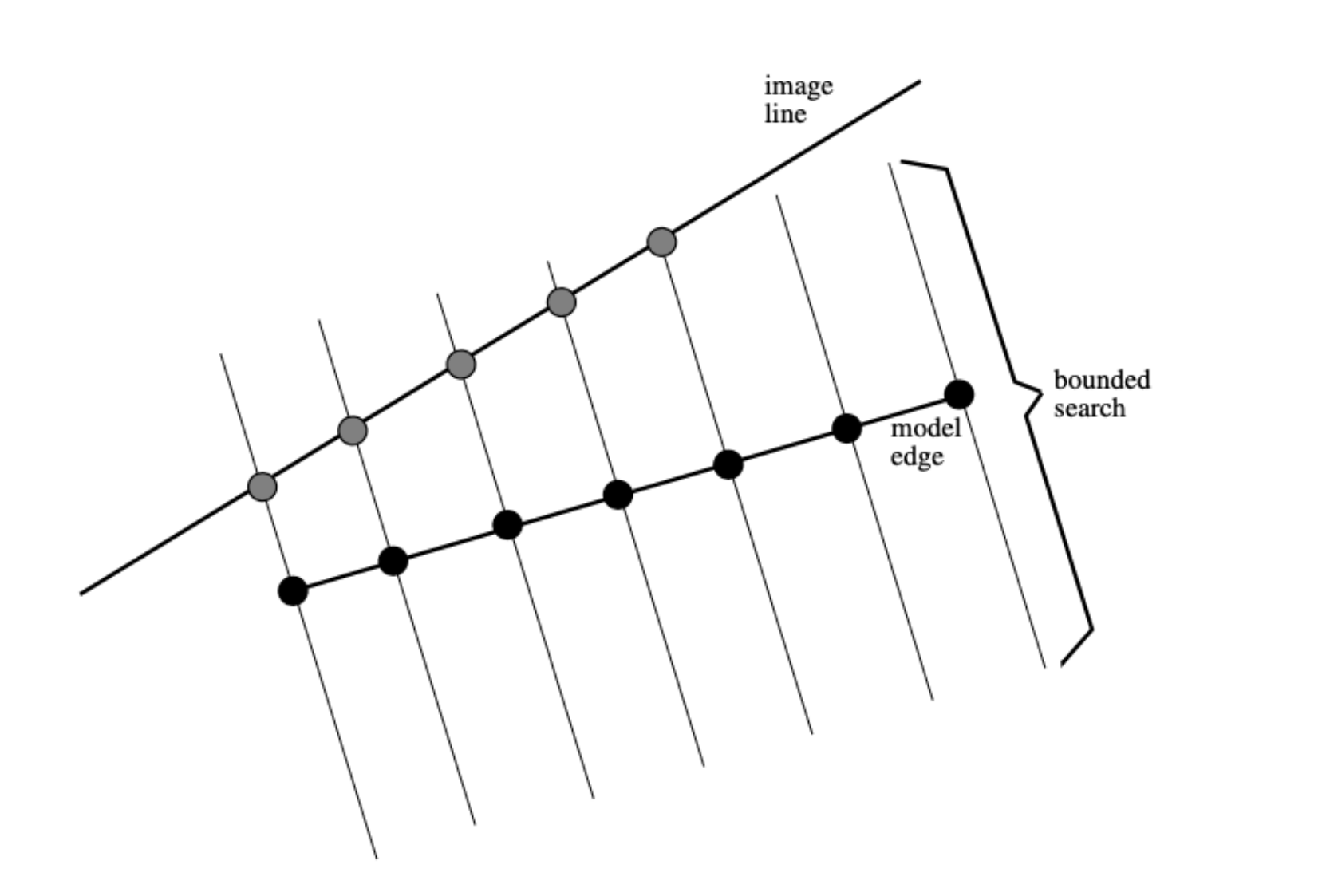}  }}
\caption{Point-to-contour (here point-to-line) correspondences are
established by searching for image points around a predicted model
point lying onto a model edge. The search is in a direction
perpendicular to the predicted model edge and is bounded.}
\label{fig:search-contour}
\end{figure}

\begin{figure*}
\centerline{
\resizebox{\textwidth}{!}{\begin{tabular}{c@{\hspace{8mm}}c}
\framebox{\includegraphics{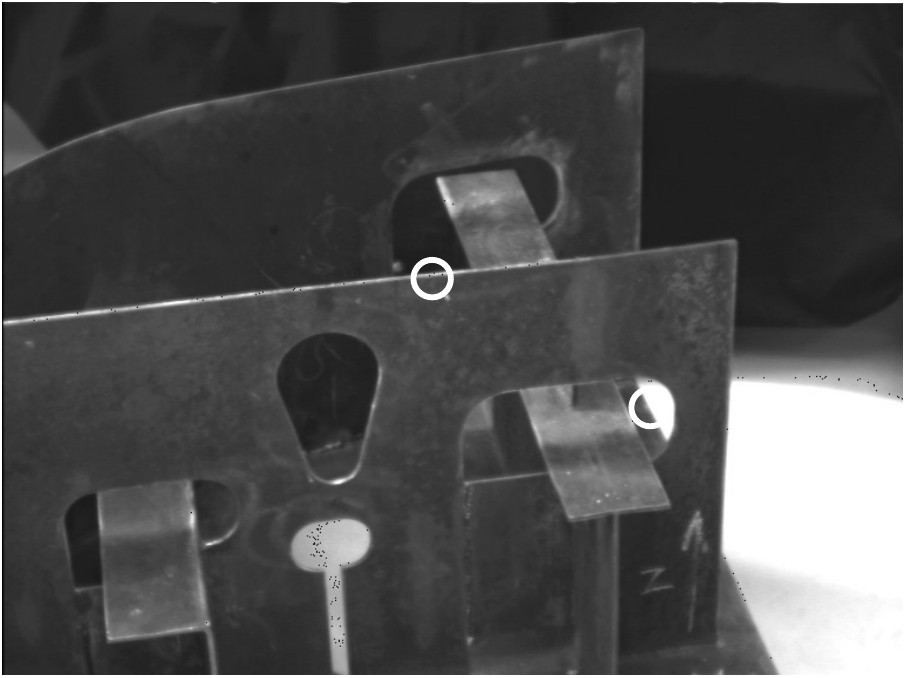}}&\framebox{\includegraphics{essai.pdf}}\\
\includegraphics[angle =-90,width=1.2\textwidth]{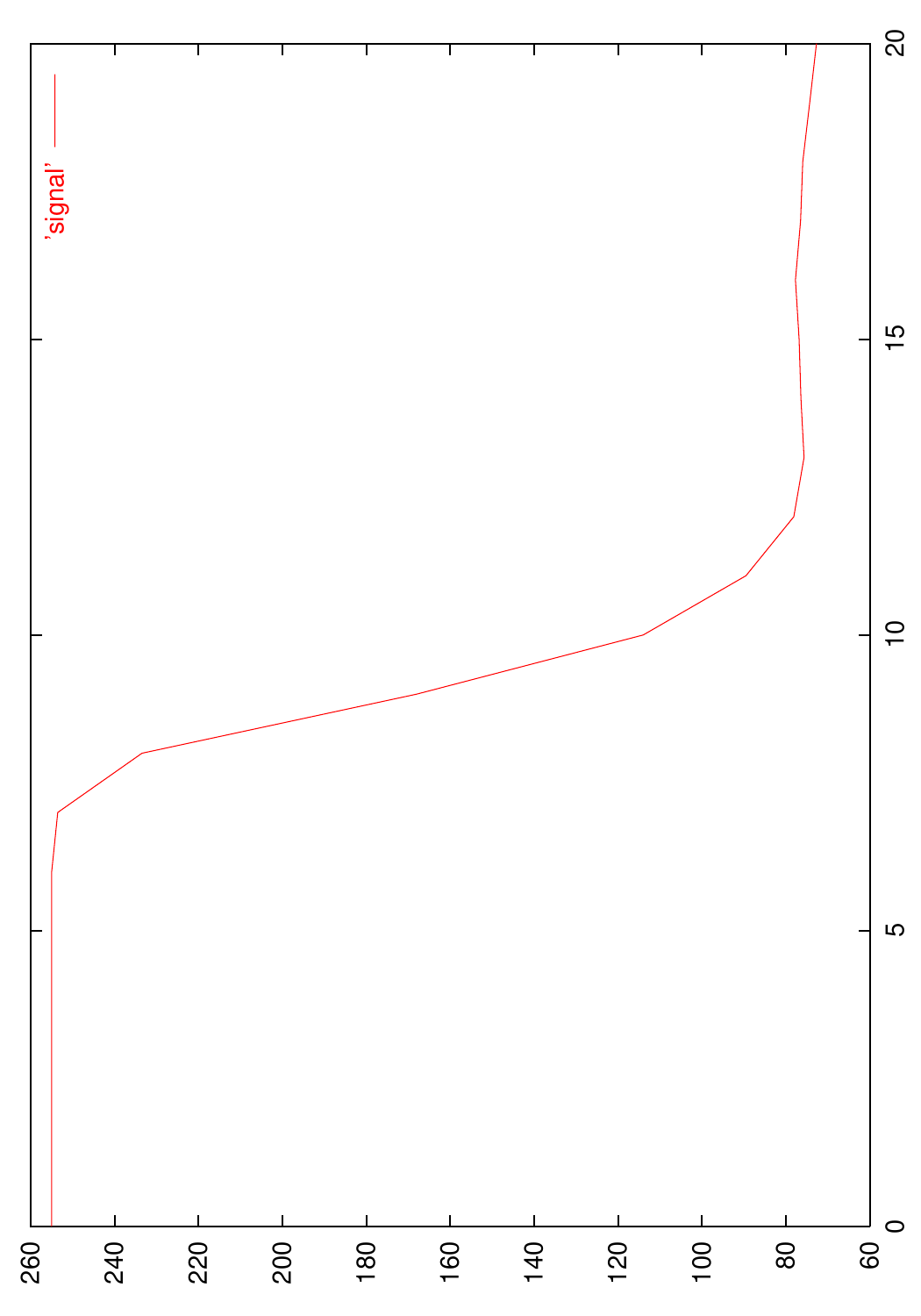}&\includegraphics[angle=-90,width=1.2\textwidth]{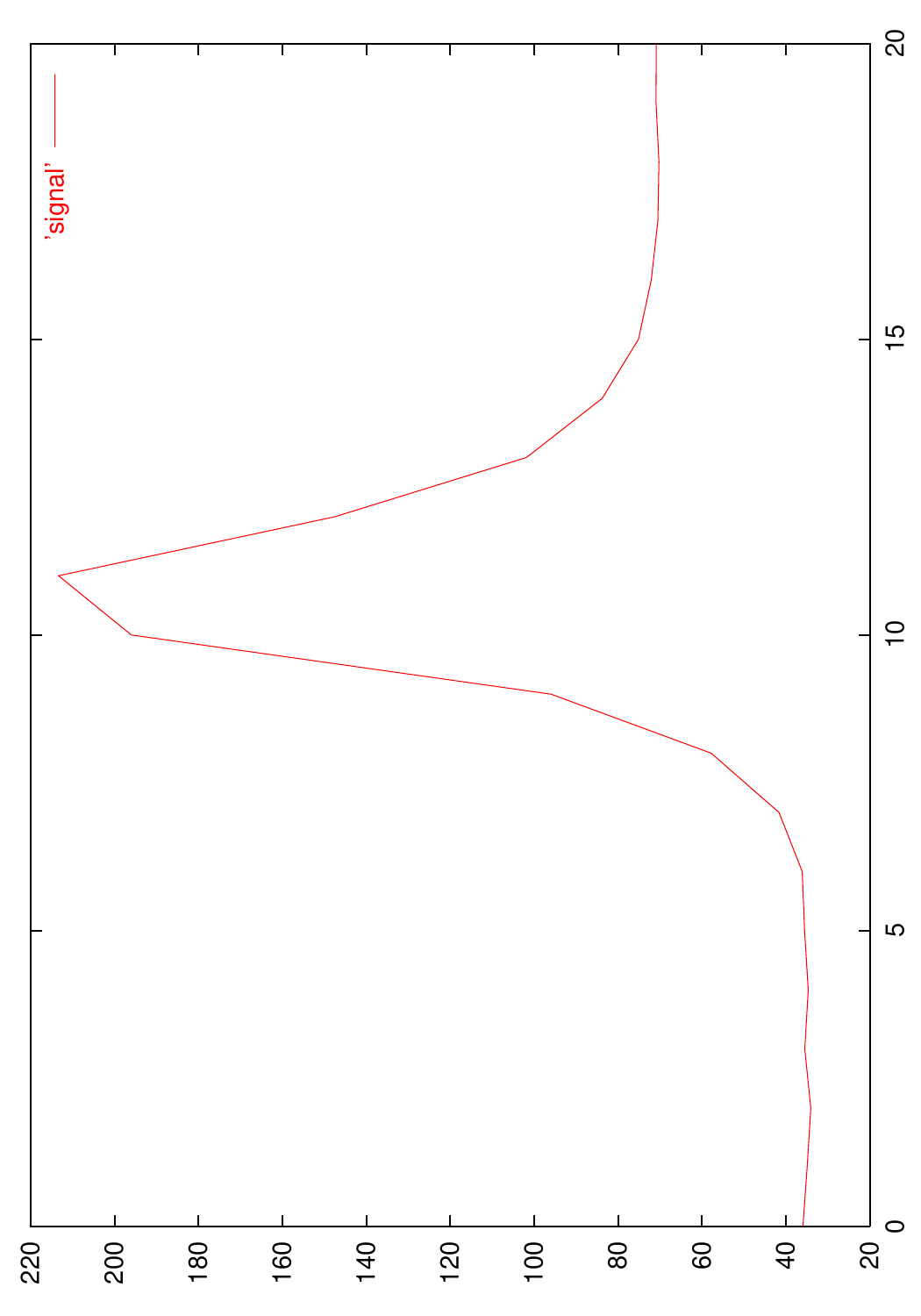}\\
\includegraphics[angle=-90,width=1.2\textwidth]{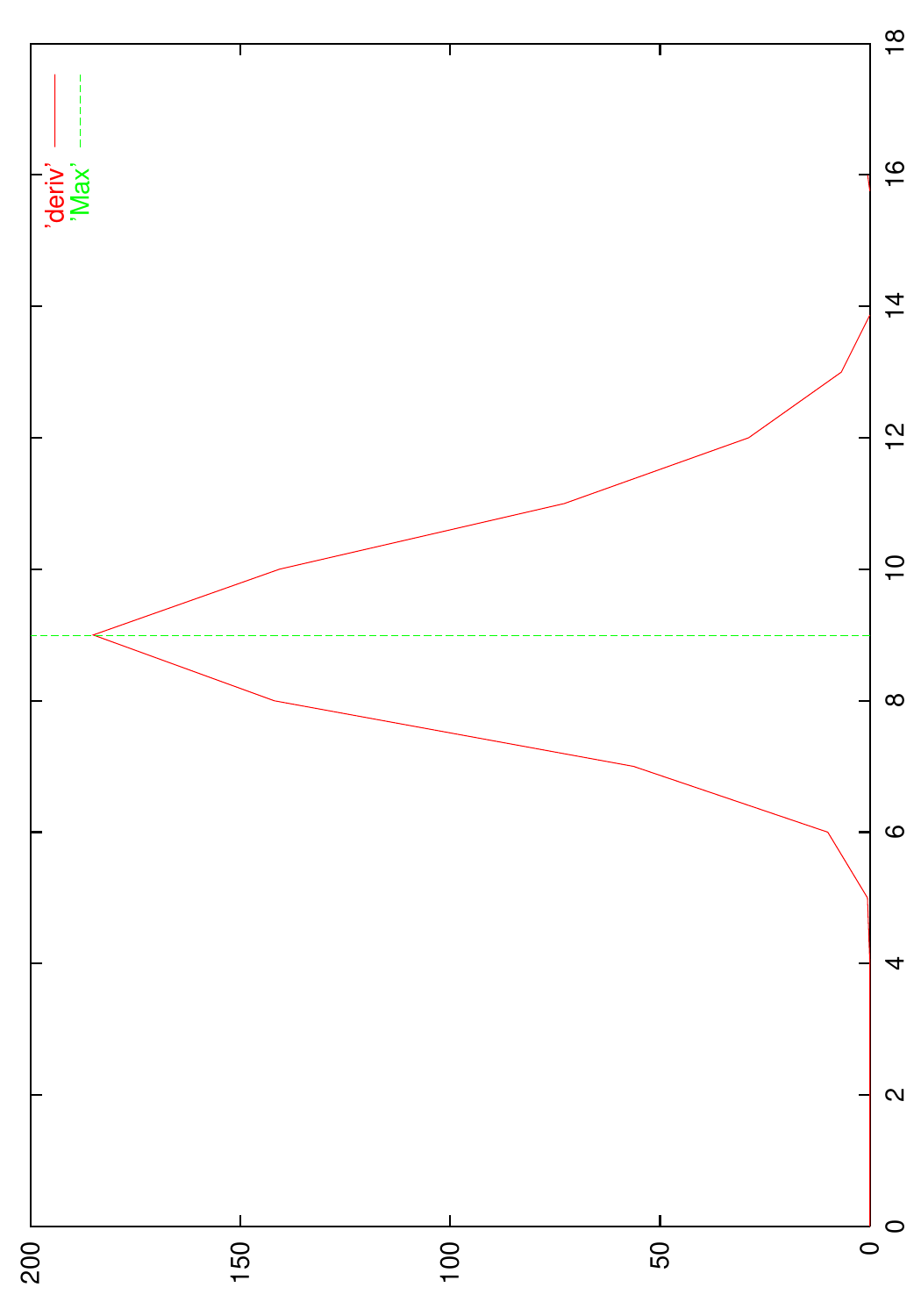}&\includegraphics[angle=-90,width=1.2\textwidth]{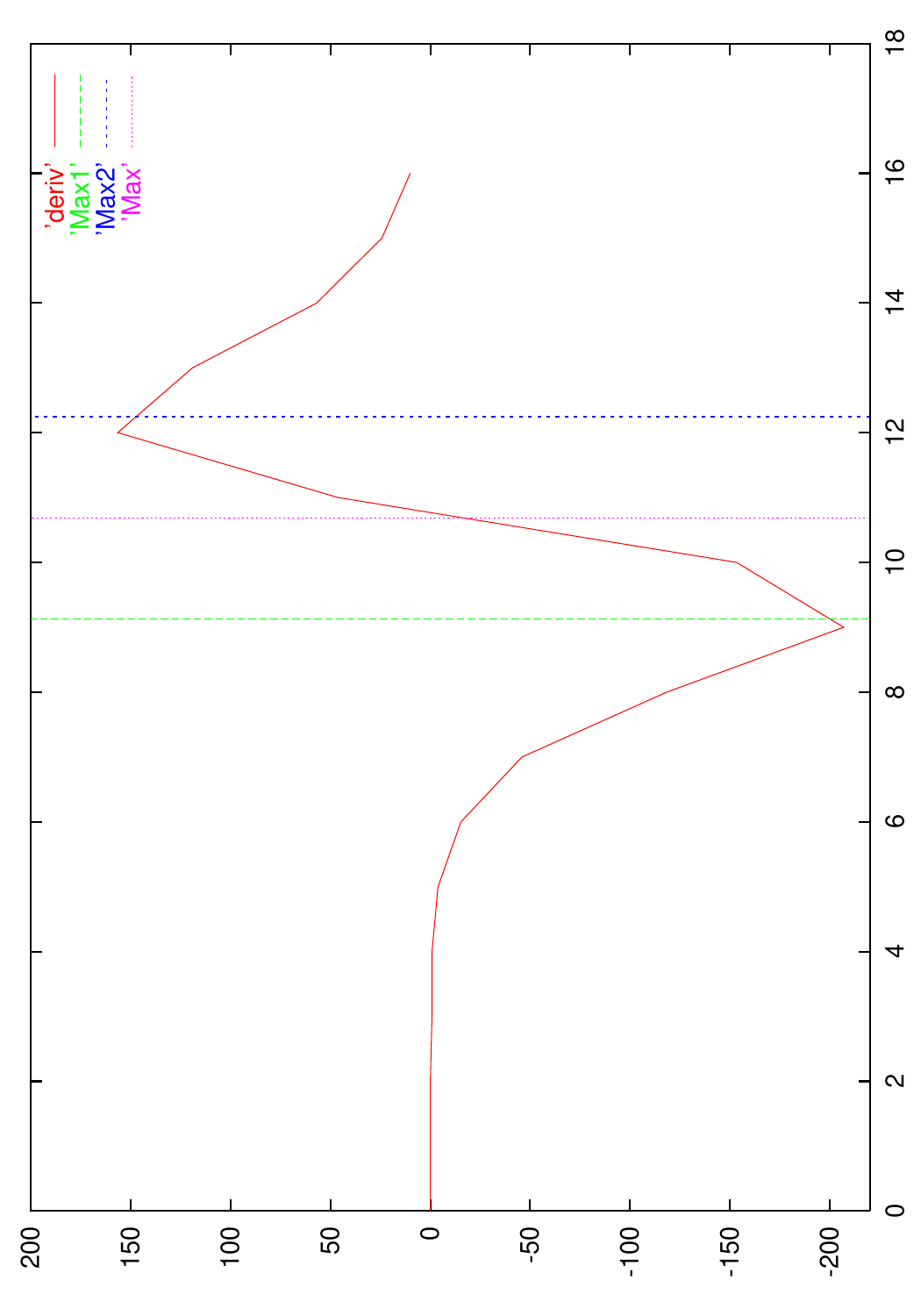}\\
\end{tabular}
}}
\caption{The data contains both step (left) and roof (right)
edges. The former are classical intensity changes. The latter
correspond to the presence of a thin plate. The intensity-profile
derivative is first computed (middle) and local maxima are located
(bottom). Whenever two local maxima are located and if their
magnitudes are similar, then a roof edge is reported and the edge is
located as a zero-crossing (bottom-right).}
\label{fig:edge-detection}
\end{figure*}

The second step is to search image contours in a direction perpendicular to the predicted model
edge, on its both sides. In practice, one-dimensional edge detection
is performed for each model point along a line perpendicular to its
supporting model edge. This one dimensional search and edge detection is
illustrated on Figures~\ref{fig:search-contour} and \ref{fig:edge-detection}.

Figure~\ref{fig:search-contour} shows a
model edge, points along this edge (black dots) and the image points
that are eventually associated with each one of these
points (grey dots). Two types of edge profiles are present in the data: step and
roof edges. Steps correspond to  classical intensity changes. Roofs
correspond to the presence in the model of thin plates. It is not
always possible to detect the contours on both sides of a thin
plate. Therefore we decided to fuse these contours into a unique
central one. A step edge is shown on
Figure~\ref{fig:edge-detection}-left and a roof edge is shown on
Figure~\ref{fig:edge-detection}-right. 

\begin{figure}
\centerline{\resizebox{0.6\textwidth}{!}{  \framebox{\includegraphics{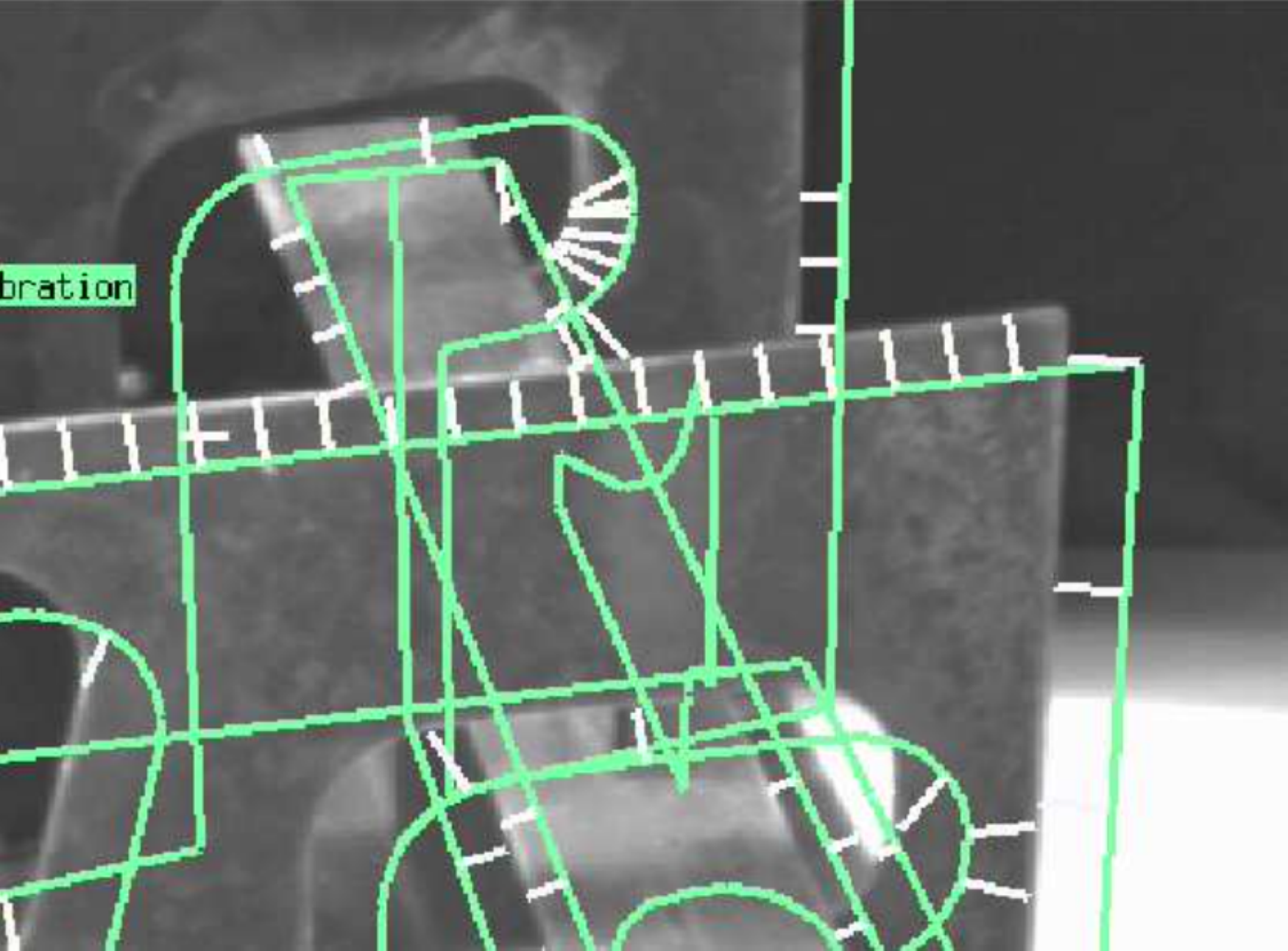}}  }}
\caption{Points along model edges are associated with image points
-- these model-contour-to-image-point correspondences are illustrated
by short segments perpendicular to the model edges. Notice that a
large number of model edges do not have image points associated with
them. The ability to restrict the number of matches to the most
reliable ones is key to successful convergence of the tracking process.}
\label{fig:data-association}
\end{figure}

The third step consists of associating an image edge-point with a
model contour. This matching is done on the basis of the
presence/absence of an edge along the direction of search within a
``short distance'' away from the predicted model contour. 
Figure~\ref{fig:data-association} shows an example. A number
of model edges are predicted in the image. The small white segments
show successful associations of model contours with image
edge points. Notice that a large number of model contours do not have
edge points associated with them either because of occlusions or because
an image edge has not been found.

\subsection{Locating a static object}

\begin{figure*}[p]
\centerline{
\resizebox{0.9\textwidth}{!}{\begin{tabular}{c@{\hspace{8mm}}c}
\framebox{\includegraphics{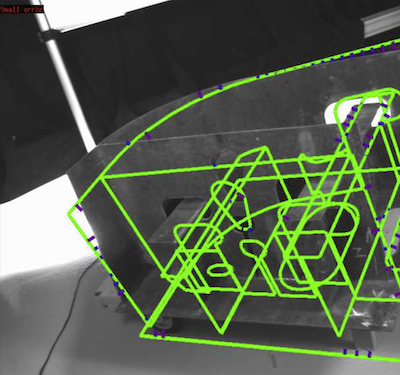}}&\framebox{\includegraphics{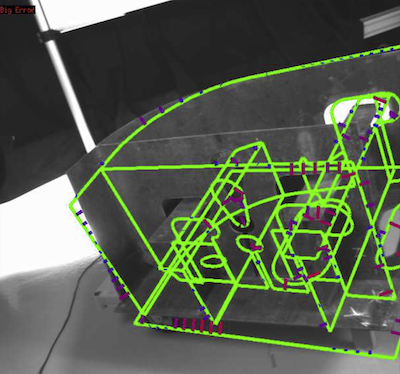}}\\
\framebox{\includegraphics{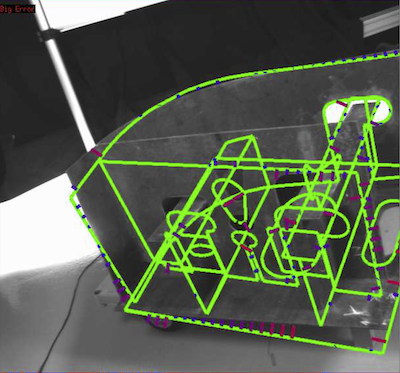}}&\framebox{\includegraphics{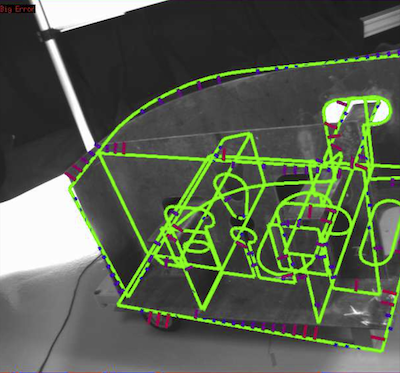}}\\
\framebox{\includegraphics{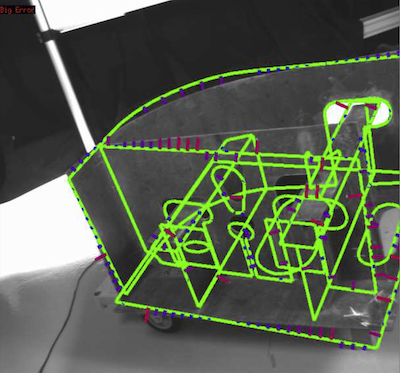}}&\framebox{\includegraphics{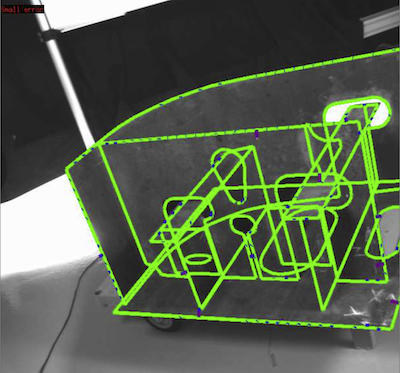}}\\
\end{tabular}
}}
\caption{This figure shows an object location example from
initialization to convergence. 18 iterations were necessary in this
case. The figure shows (left to right and top to down) the
initialization, as well as iterations 3, 6, 9, 12, and 18. Although
the algorithm performs hidden-line elimination and considers only the
model points predicted visible, the on-line display shows all (visible and
hidden) model edges.}
\label{fig:tracking-pose}
\end{figure*}

The first experiment consists of using one camera with fixed 
internal parameters observing a static object.
A complete object location sequence is shown on
Figure~\ref{fig:tracking-pose}. The initial model location was set
manually. After 18 iterations the pose parameters converged to a
stable solution corresponding to an average image error of 0.8
pixels. 

Each iteration comprises (i)~the gathering of a new image, (ii)~model
prediction in the image using the previously estimated pose
parameters, (iii)~hidden-line elimination, (iv)~detecting an image contour
match for each model point, and (v)~one Gauss-Newton iteration based on
valid point-to-contour correspondences. Since with one camera the
algorithm runs and 25Hz, it takes 0.7 seconds to reach convergence in
this case. 
Once convergence is reached, the system is ready for tracking.

\subsection{Tracking the focal length}

\begin{figure*}[p]
\centerline{
\resizebox{0.9\textwidth}{!}{\begin{tabular}{c@{\hspace{8mm}}c}
\framebox{\includegraphics{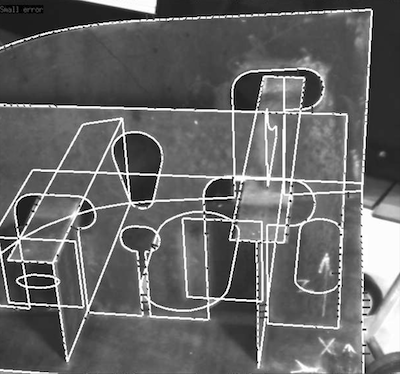}}&\framebox{\includegraphics{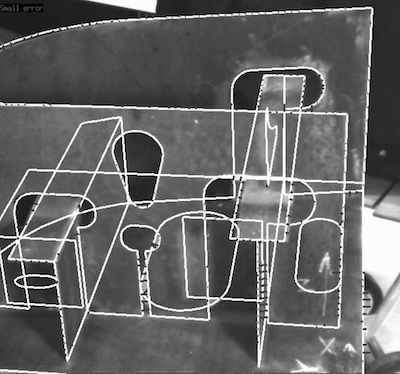}}\\
\framebox{\includegraphics{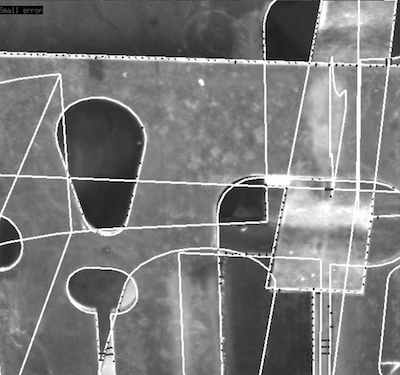}}&\framebox{\includegraphics{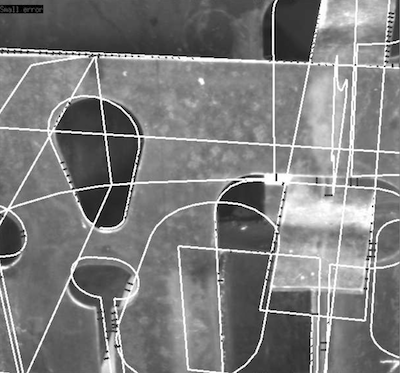}}\\
\framebox{\includegraphics{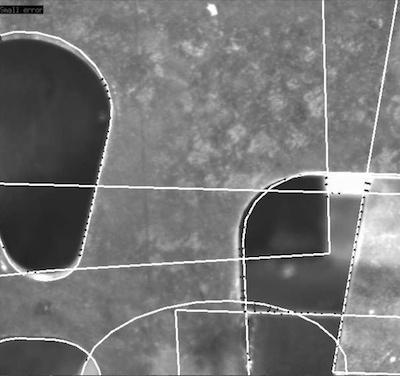}}&\framebox{\includegraphics{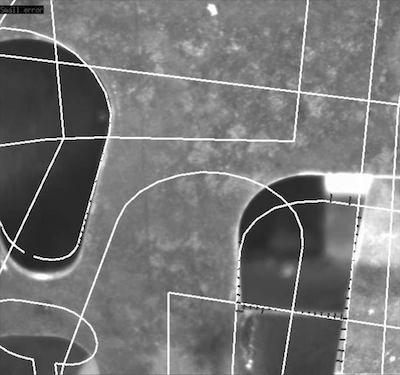}}\\
\end{tabular}
}}
\caption{The effect of zooming in and zooming out can be taken into
account by the tracker (left column). The right column shows the
effect of tracking with fixed focal length: the zoom effect is
compensated by variations in depth, but this compensation is far from
being perfect.}
\label{fig:tracking-focal}
\end{figure*}

The second experiment consists of using one camera with varying focal
length observing a static object. The focal length varies back and
forth from its minimal to its maximal values with a ratio of 1:3. The
left column of Figure~\ref{fig:tracking-focal} shows the result of
tracking in such a case. Both the pose and the focal length are
allowed to vary during the tracking process using the adequate
Jacobian matrix, i.e., sections~\ref{subsection:varying-focal-1} and
\ref{section:varying-focal-2}.

The right column of Figure~\ref{fig:tracking-focal} shows the same
experimental setup but the focal length is set to a fixed value. The
tracker tries to compensate the zoom effect with depth variations
but, clearly, the results are not very good. 

Table~\ref{table:zoom-tracking}
summarizes a series of tests. The first test (top) is
designed to track the focal length as just described. Small variations
in depth are du to the fact that the optical center moves while the
zoom varies. The second test (middle) maintains the focal lenght to
its initial value and therfore the depth compensate for zoom
variations. The third test (bottom) allows both internal and external
parameters to vary. Although the optical center moves a lot, the focal
lentgth values estimated in this case are very close to those
estimated in the first test.

This is clear experimental evidence that the method described in this
paper is able to make a clear distinction between depth variations and
zoom variations. The only case that confuses the tracker is when the
object is planar and lies in a fronto-parallel plane. 

\begin{table}
\begin{center}
\begin{tabular}{|c|c|c|c|c|}
\hline
                & $f (pixels)$  & $u_0$ (pixels) & $v_0$ (pixels) & Depth (mm) \\
\hline
Focal length \& & 1615.2  & 288.4 & 386.8 & 2351.7 \\
External        & 3226.4  & 288.4 & 386.8 & 2503.7 \\
Parameters      & 4899.5  & 288.4 & 386.8 & 2553.4 \\
\hline
External        & 1569.8  & 289.0 & 386.3 & 2282.1 \\
Parameters      & 1569.8  & 289.0 & 386.3 & 1293.8 \\
                & 1569.8  & 289.0 & 386.3 & 840.2  \\
\hline
All             & 1566.7  & 288.4 & 386.8 & 2476.8 \\
Parameters      & 3190.2  & 335.5 & 623.7 & 2512.3 \\
                & 4956.2  & 164.4 & 105.2 & 2527.5 \\
\hline
\end{tabular}
\end{center}
\caption{This table shows a series of three tests where only the focal
length varies in practice.}
\label{table:zoom-tracking}
\end{table}

\subsection{Camera calibration during tracking}

Modern digital cameras have built-in auto-focus. The ability to track
an object with such a camera necessitates on-line estimation of
internal camera parameters because a change in focus is associated
with changes in internal camera parameters. 

The third experiment consists of estimating the internal camera
parameters -- tracking with one camera and 9 parameters while the
object is allowed to move. Starting with a good initial guess and
after only a few iterations, the tracker
converges such that the estimated camera parameters match the real
values. 

In order to assess the quality of the estimated camera parameters we
compared the results of tracking with the result obtained by using a
calibration jig, Table \ref{table:calibration}. The first row displays
parameter values determined off-line using a classical camera
calibration technique in conjunction with a calibrated object (the
accuracy of the 3-D measurements are of about 0.01mm). The subsequent
rows correspond to experiments performed with the tracker. Each
experiment consists of 100 measurements. Each experiment correspond to
a different setup -- a different relative position between object and camera.

The discrepancies between the two sets of
parameters appear from the fact that the ship part is a welded object
which does not perfectly correspond to its CAD model. The accuracy of
the ship part real size with respect to its theoretical CAD
representation is of the order of a few milimeters.

\begin{table}
\begin{center}
\begin{tabular}{|c|c|c|c|c|c|c|}
\hline
  & $f_u$  & std &  $u_0$  &  std & $v_0$ & std  \\
\hline
off-line & 1524.5 &  &    283.1 & &  400.2 &  \\
\hline
% il y a un decalage de 2, l'experimentation 1 est la 3
Exp. 1  & 1468.7  & 3.1  &  282.8 & 1.6   & 376.6 & 1.2 \\
Exp. 2  & 1479.6  & 4.2  &  321.7 & 2.2   & 349.1 & 3.2 \\
Exp. 3  & 1490.1  & 2.65 &  317.2 & 2.1   & 347.9 & 2.6 \\
Exp. 4  & 1458.4  & 3.1  &  263.2 & 1.4   & 366.1 & 1.6 \\
Exp. 5  & 1433.3  & 3.24 &  254.1 & 3.91  & 380.5 & 1.94 \\
Exp. 6 & 1554.3  & 0.83 &  286.0 & 1.50  & 338.9 & 0.76 \\
\hline
\end{tabular}
\end{center}
\caption{\label{table:calibration} The internal camera parameters obtained with classical
off-line calibration (first row) are compared with those obtained during
the tracking process.}
\end{table}

\subsection{Tracking with three cameras}

The last experiment that we describe uses 3 static cameras observing a
moving object. With three cameras the tracking system runs at 12Hz
thus allowing objects to freely move in the viewing volume. Notice
that the camera setup is arbitrary, no common field of view in between
any camera pair is necessary. One camera has a panoramic field of view
while the other two cameras zoom onto various object parts.

The camera system is calibrated using the previously described
experiments for each individual camera. During this experiment the
internal camera parameters associated with all three cameras are fixed
as well as the relative positions and orientations of the
cameras. Therefore the Jacobian used by the tracker is given by
eq.~\ref{eq:multiple-Jacobian-2}. From the Jacobian singularity
analysis carried out above it is clear that such a 3 camera tracker
has no singularity associated with it. The minimal data association
required by the tracker is composed of 6 model points matching 3 image
contours, with two points matching one contour. 

In practice, the tracker has to deal with partial occlusions in all
images, global occlusion in one or two images, deficiency of one or
two cameras, etc. The advantage of the 3 camera system is that the
minimal matching requirement just described may be distributed. The
tracker is able perform well with as little matches as two colinear model
points visible in each one of the 3 images.

\begin{figure}
\centerline{
\resizebox{\textwidth}{!}{\begin{tabular}{c@{\hspace{8mm}}c@{\hspace{8mm}}c}
\framebox{\includegraphics{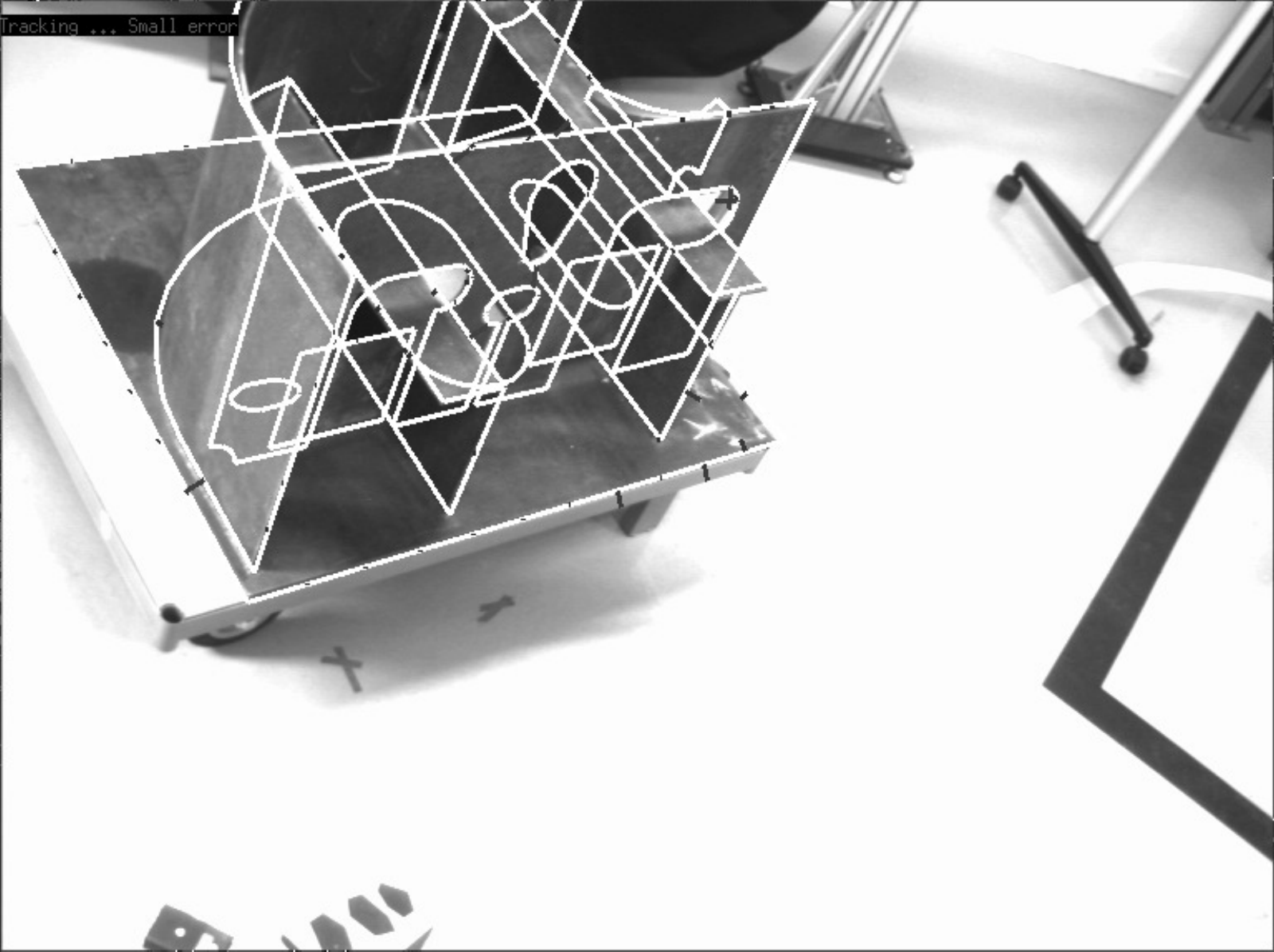}}&\framebox{\includegraphics{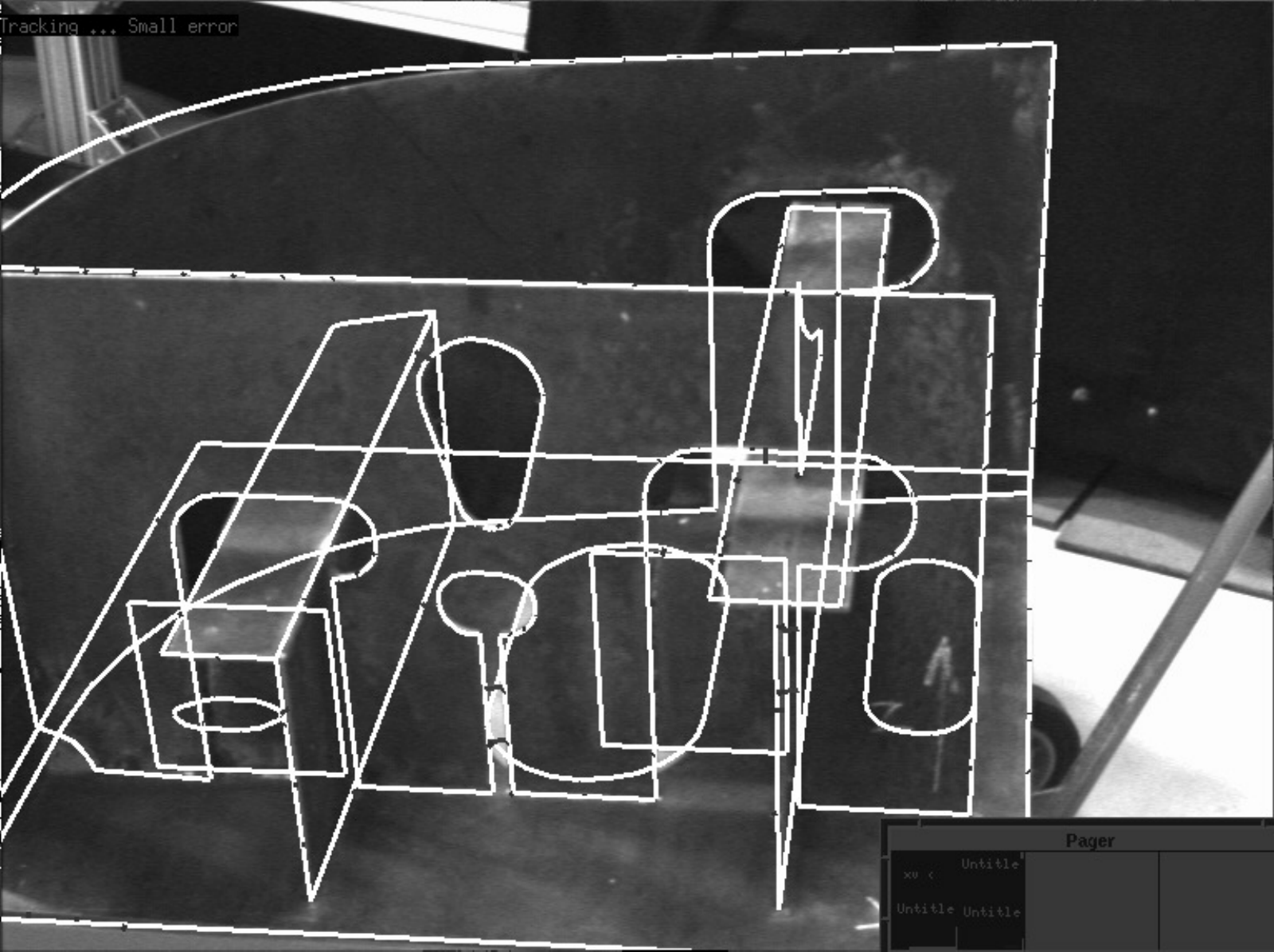}}&\framebox{\includegraphics{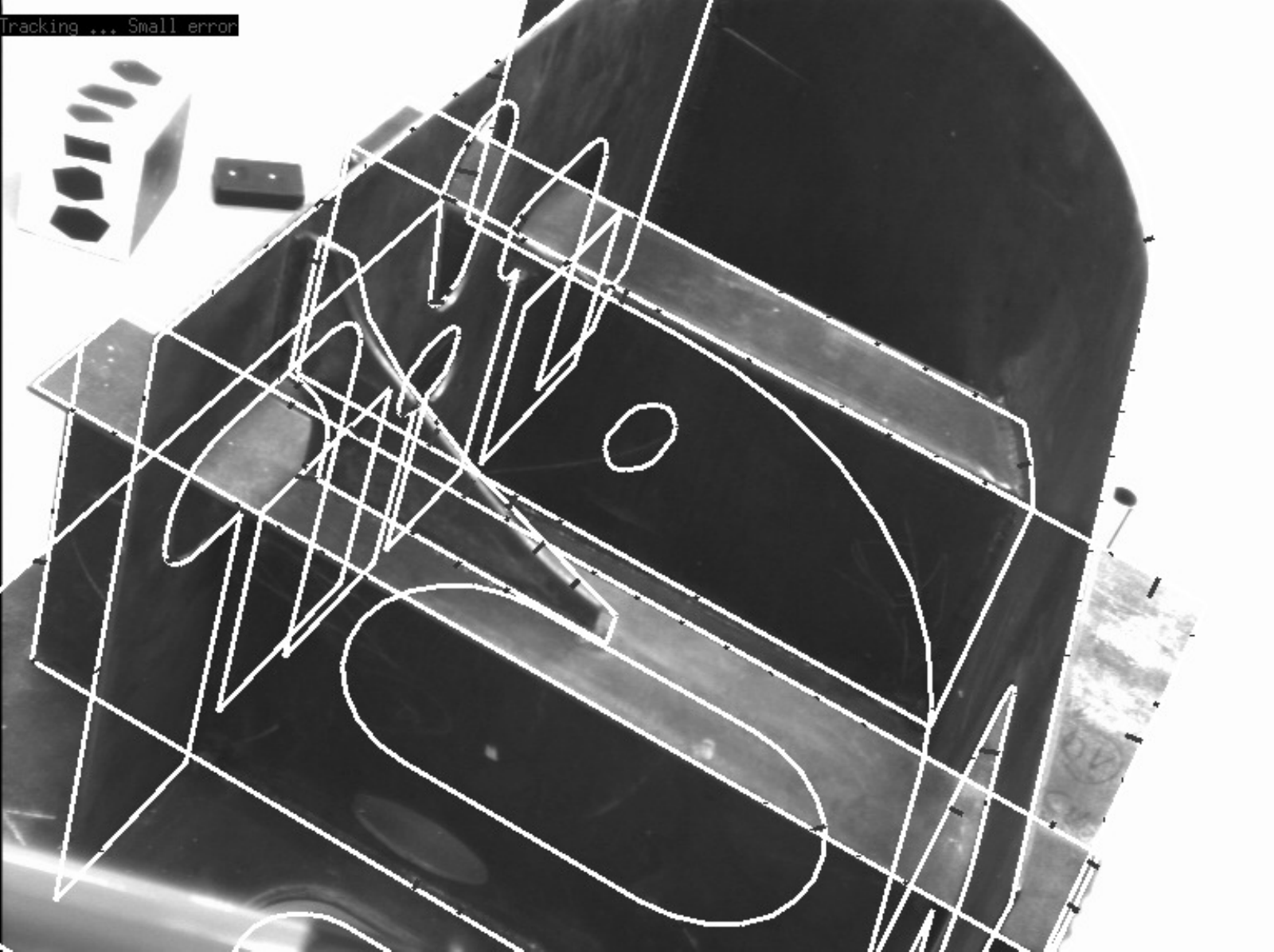}}\\
\framebox{\includegraphics{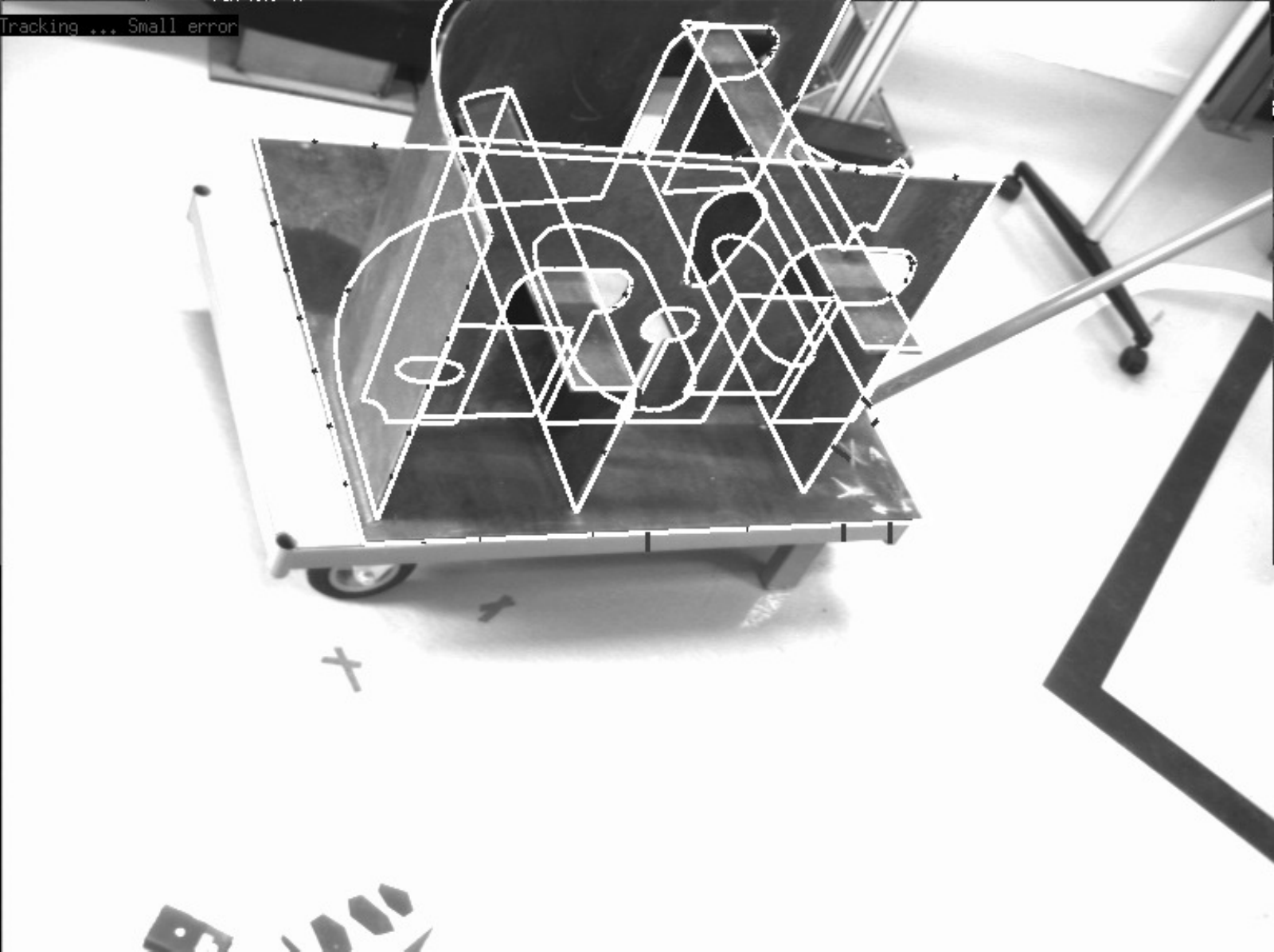}}&\framebox{\includegraphics{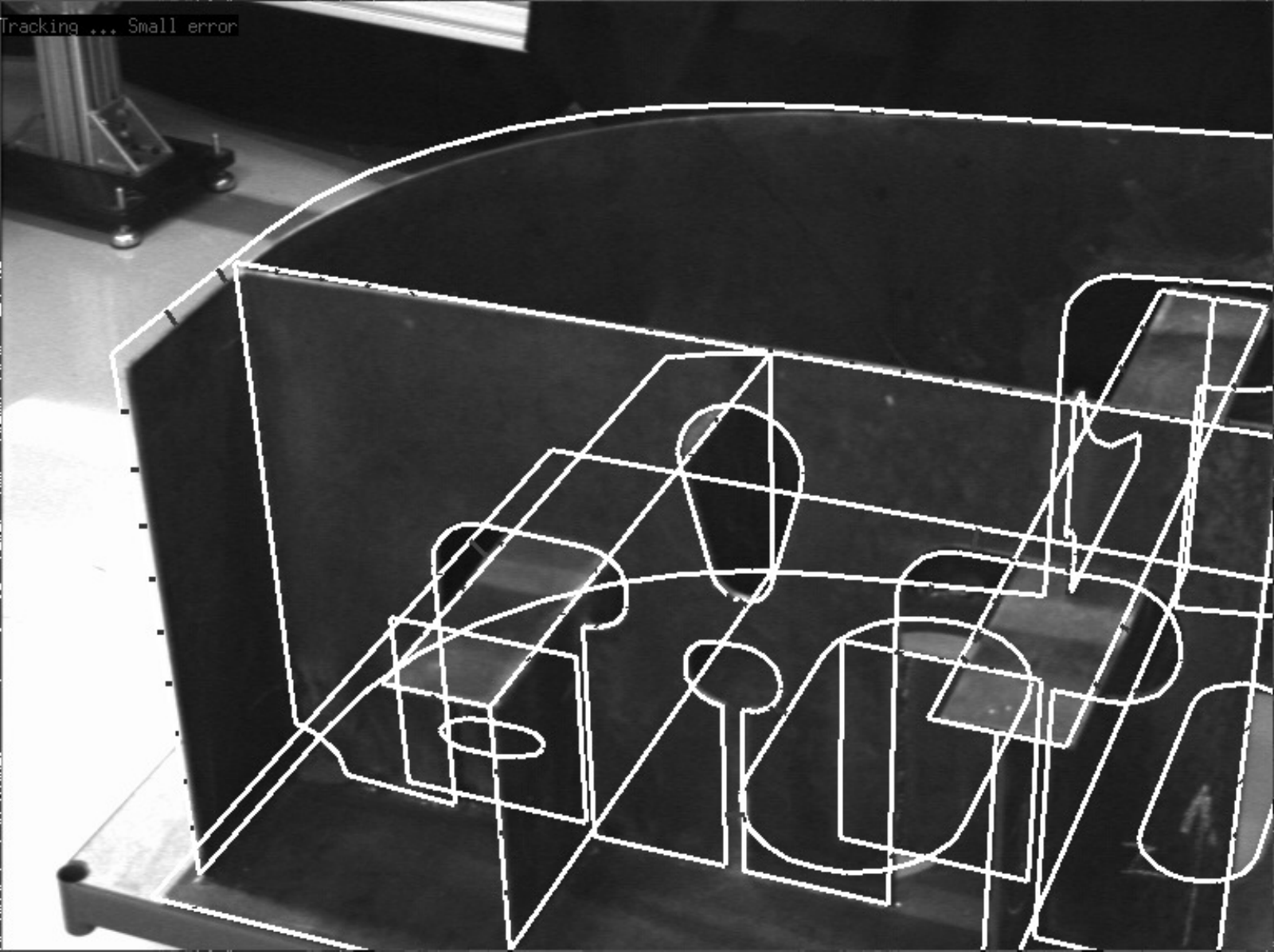}}&\framebox{\includegraphics{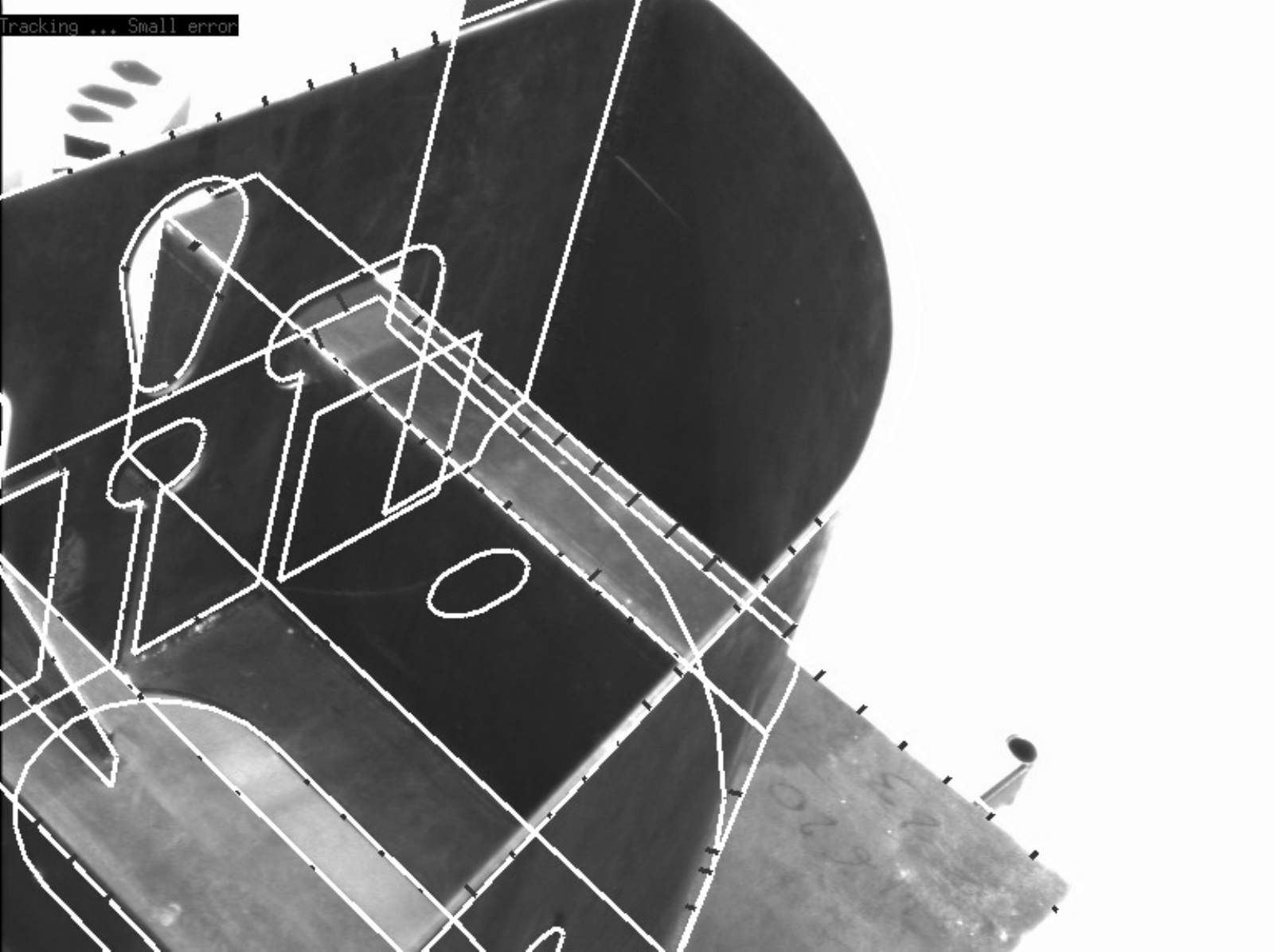}}\\
\framebox{\includegraphics{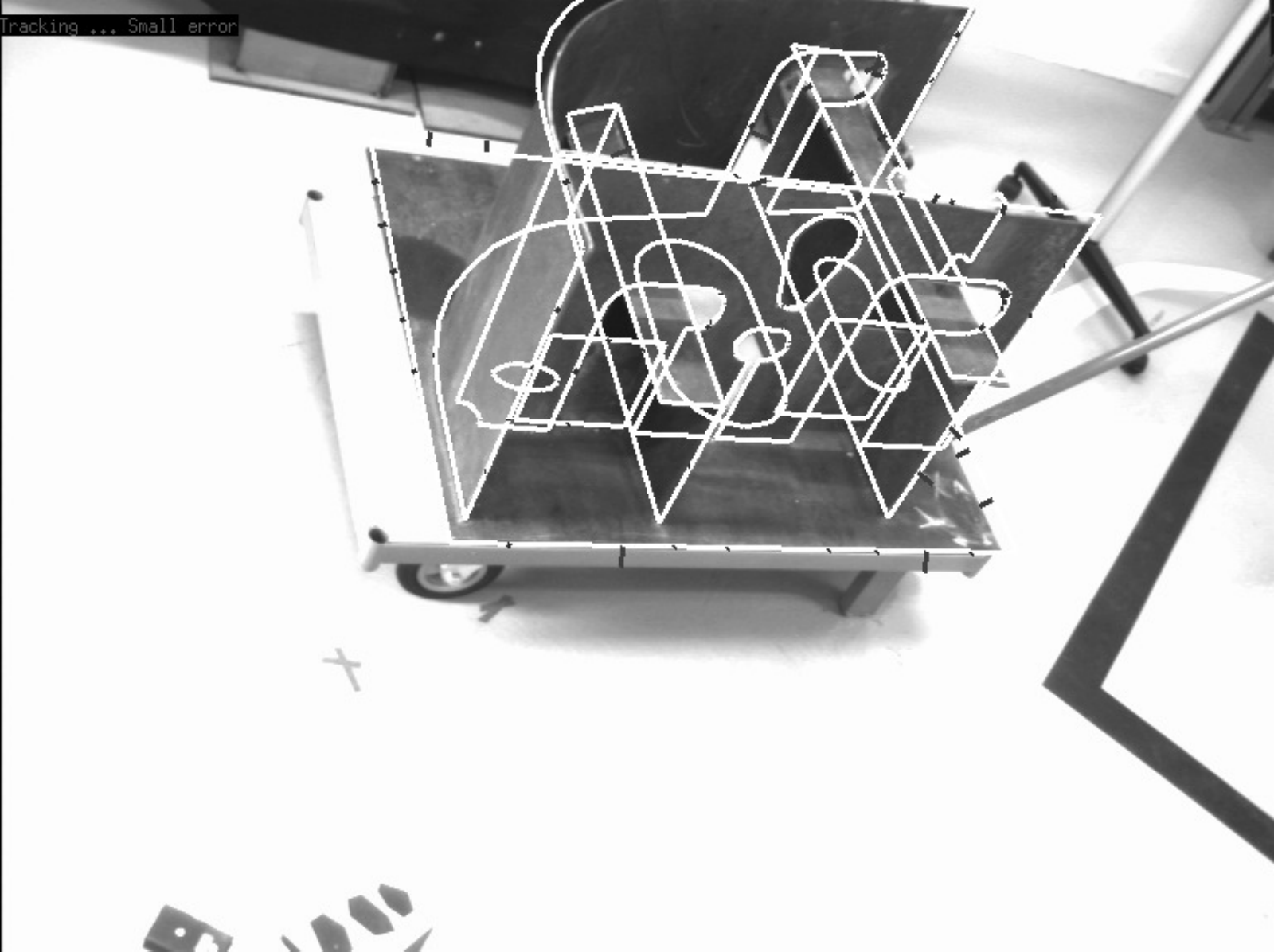}}&\framebox{\includegraphics{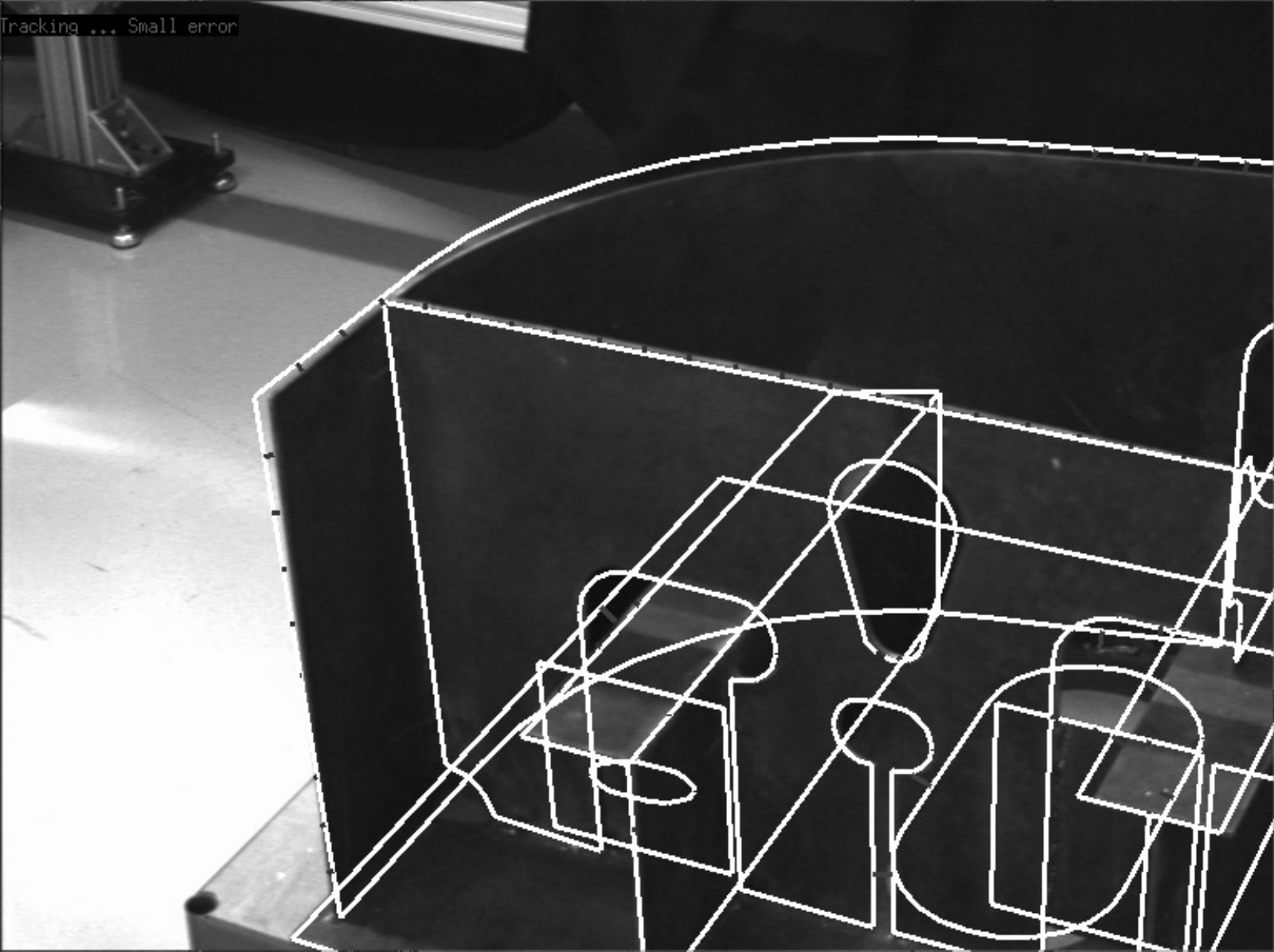}}&\framebox{\includegraphics{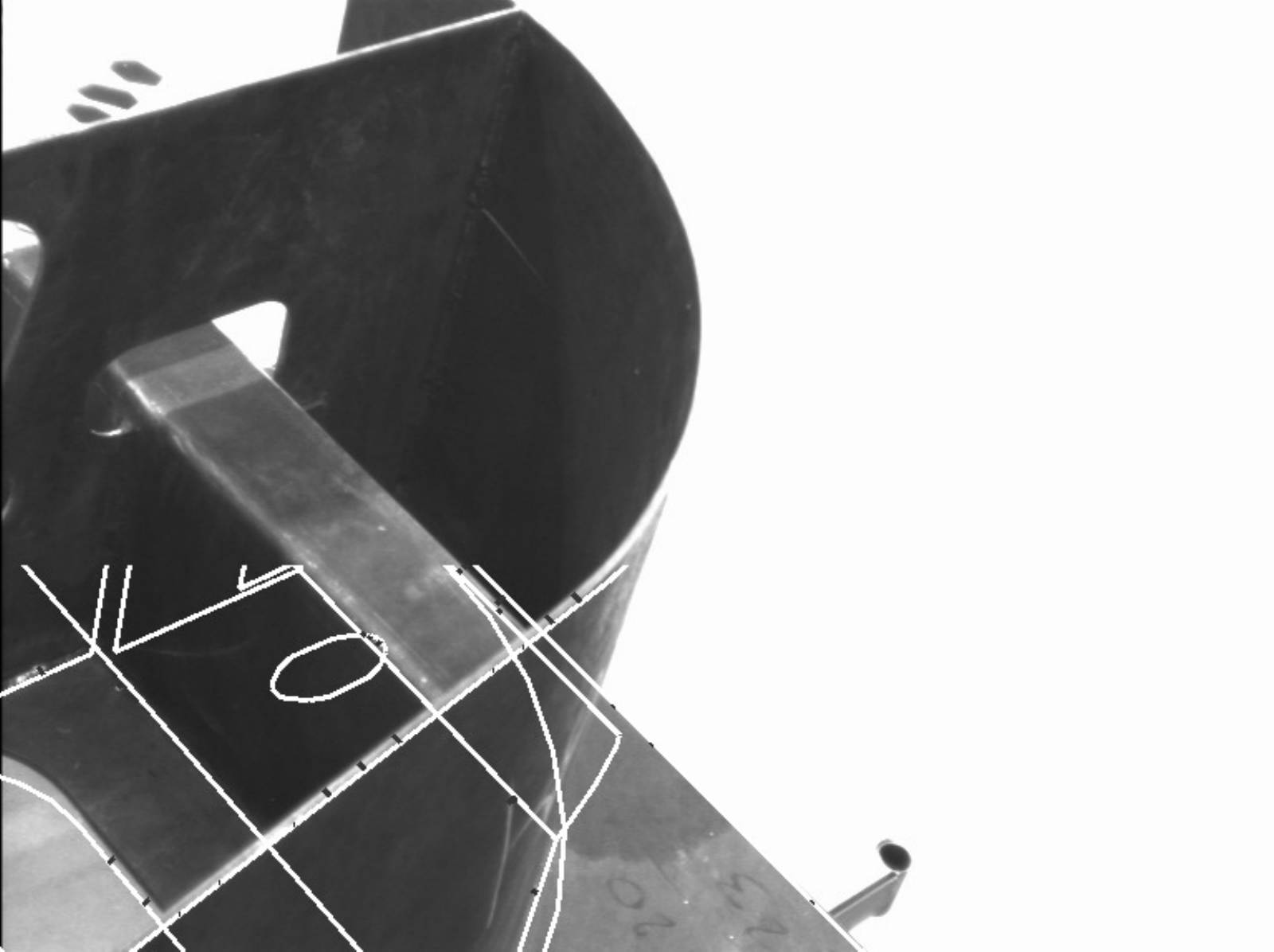}}\\
\framebox{\includegraphics{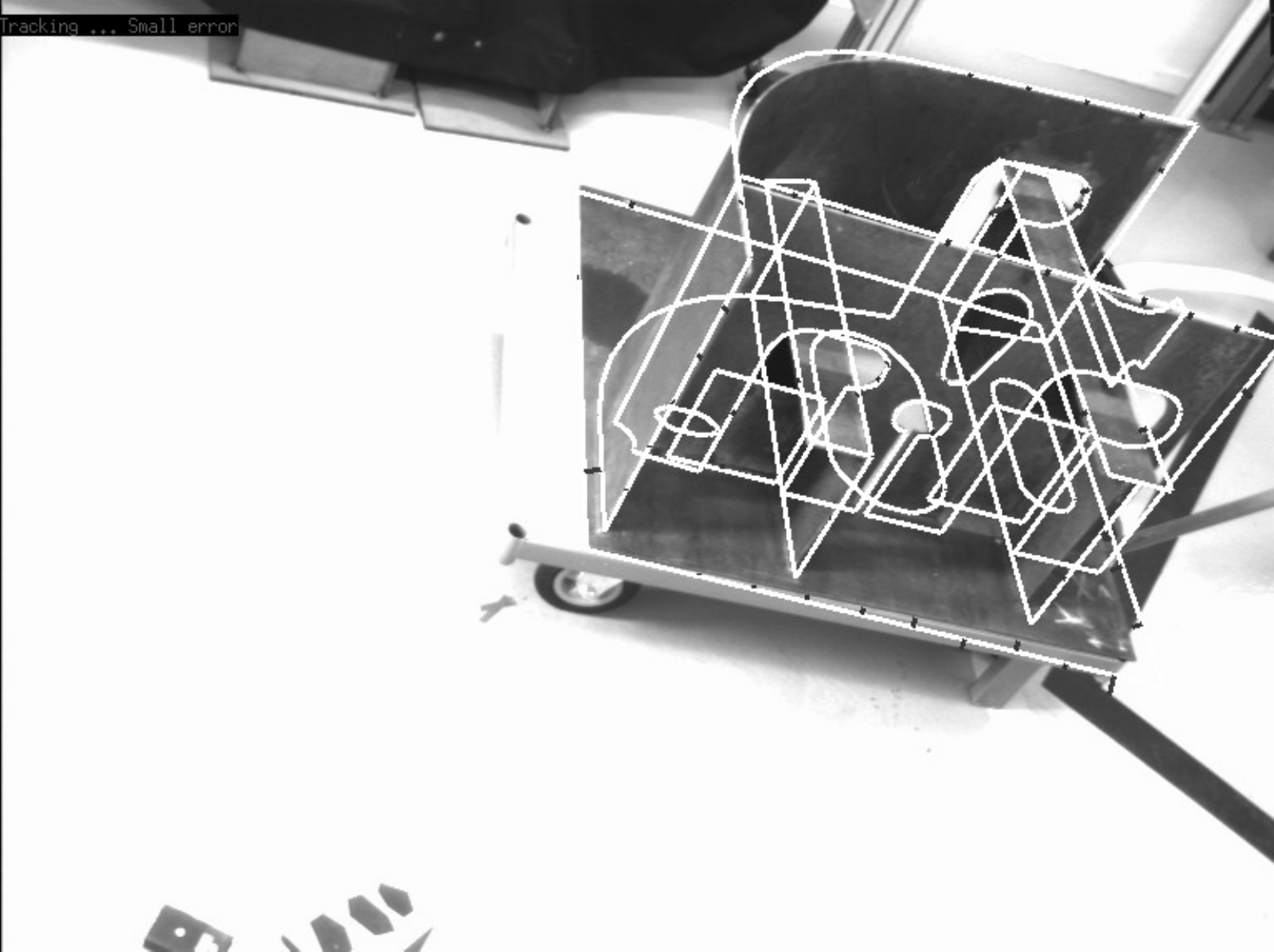}}&\framebox{\includegraphics{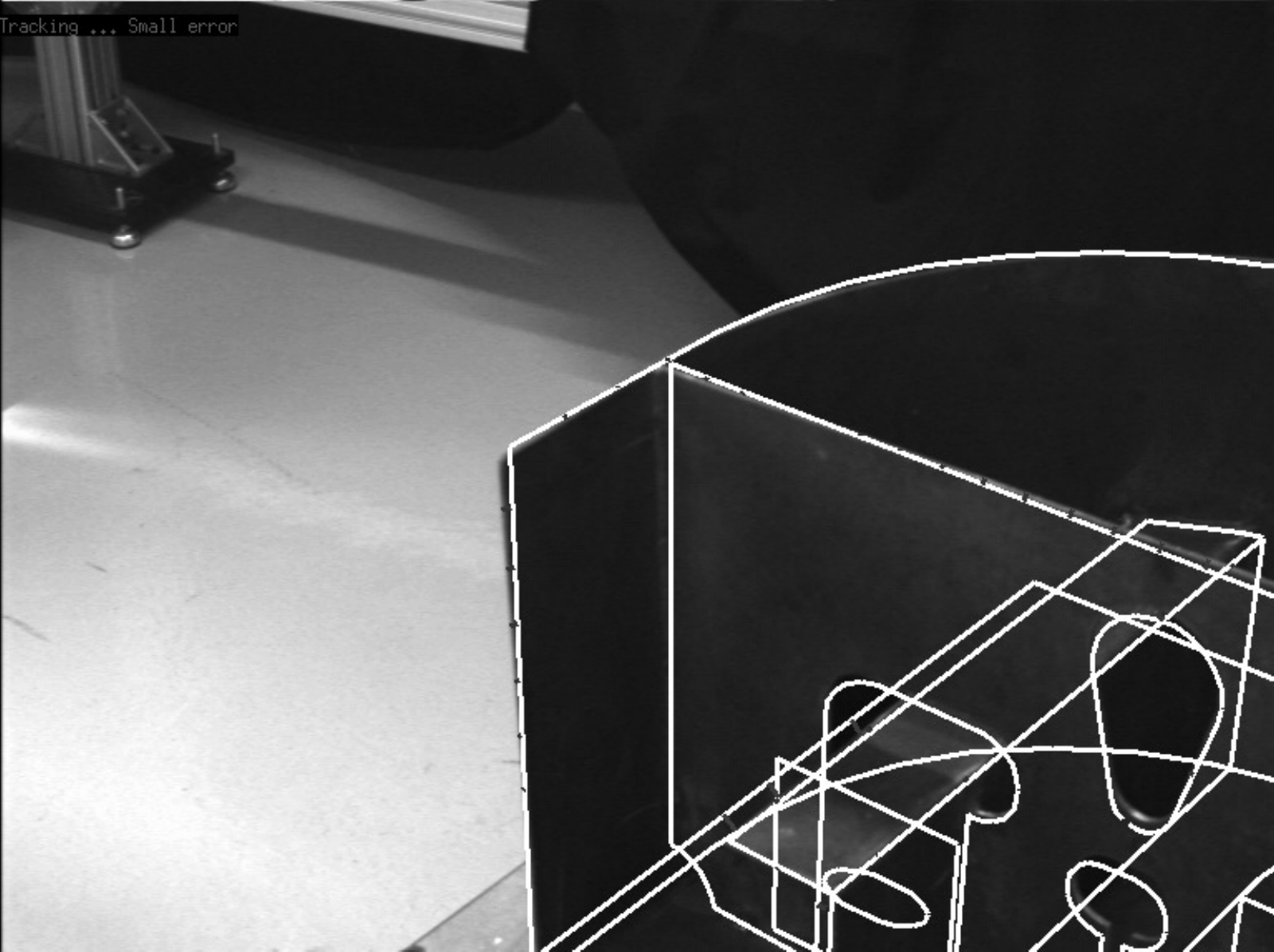}}&\framebox{\includegraphics{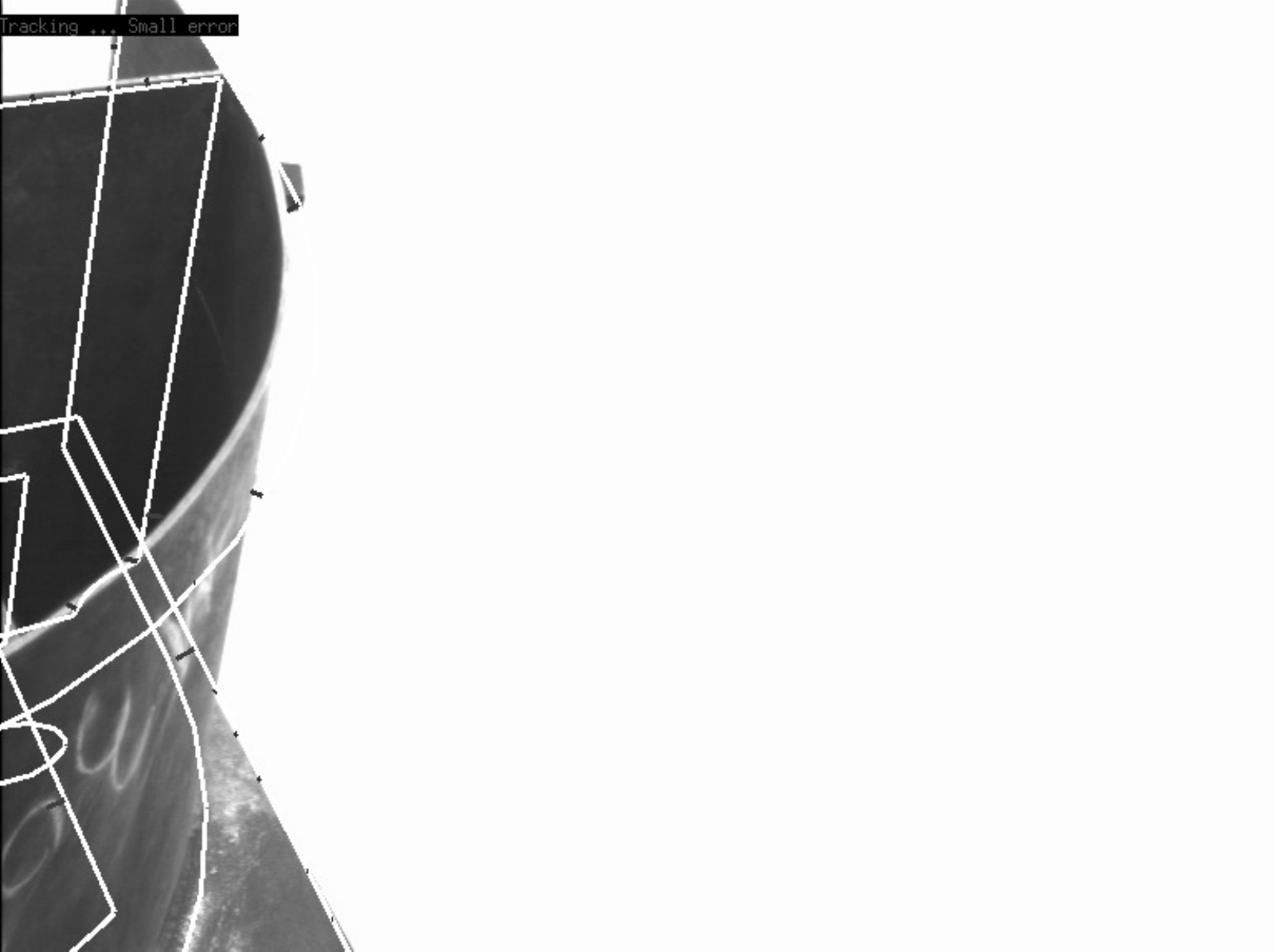}}\\
\end{tabular}
}}
\caption{A tracking sequence. The object rotates and translates onto
the floor. Notice that the three cameras have very different points of 
view.}
\label{fig:tracking-3cameras}
\end{figure}

\begin{figure}
\centerline{
\resizebox{\textwidth}{!}{\begin{tabular}{c@{\hspace{8mm}}c@{\hspace{8mm}}c}
\framebox{\includegraphics{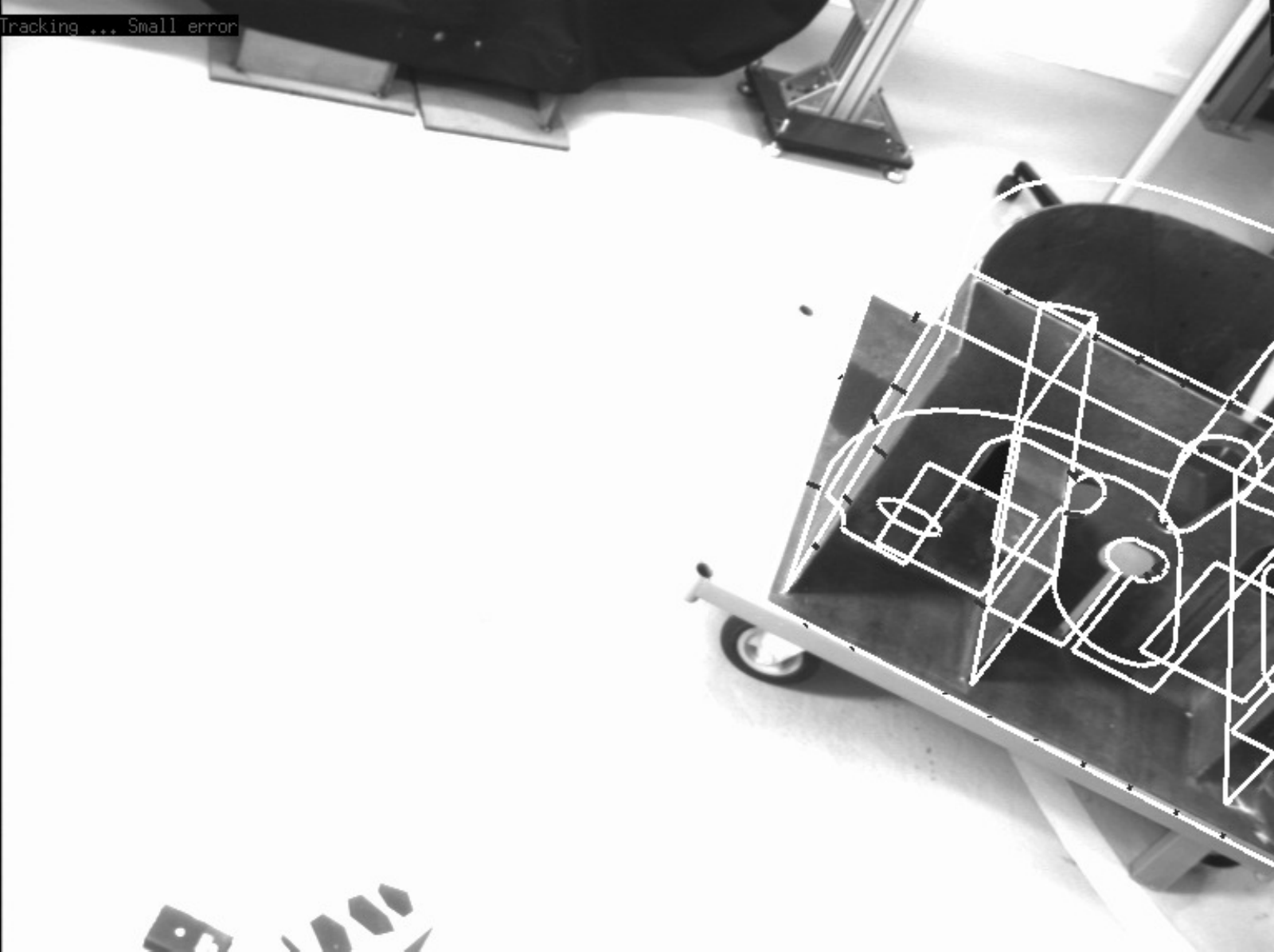}}&\framebox{\includegraphics{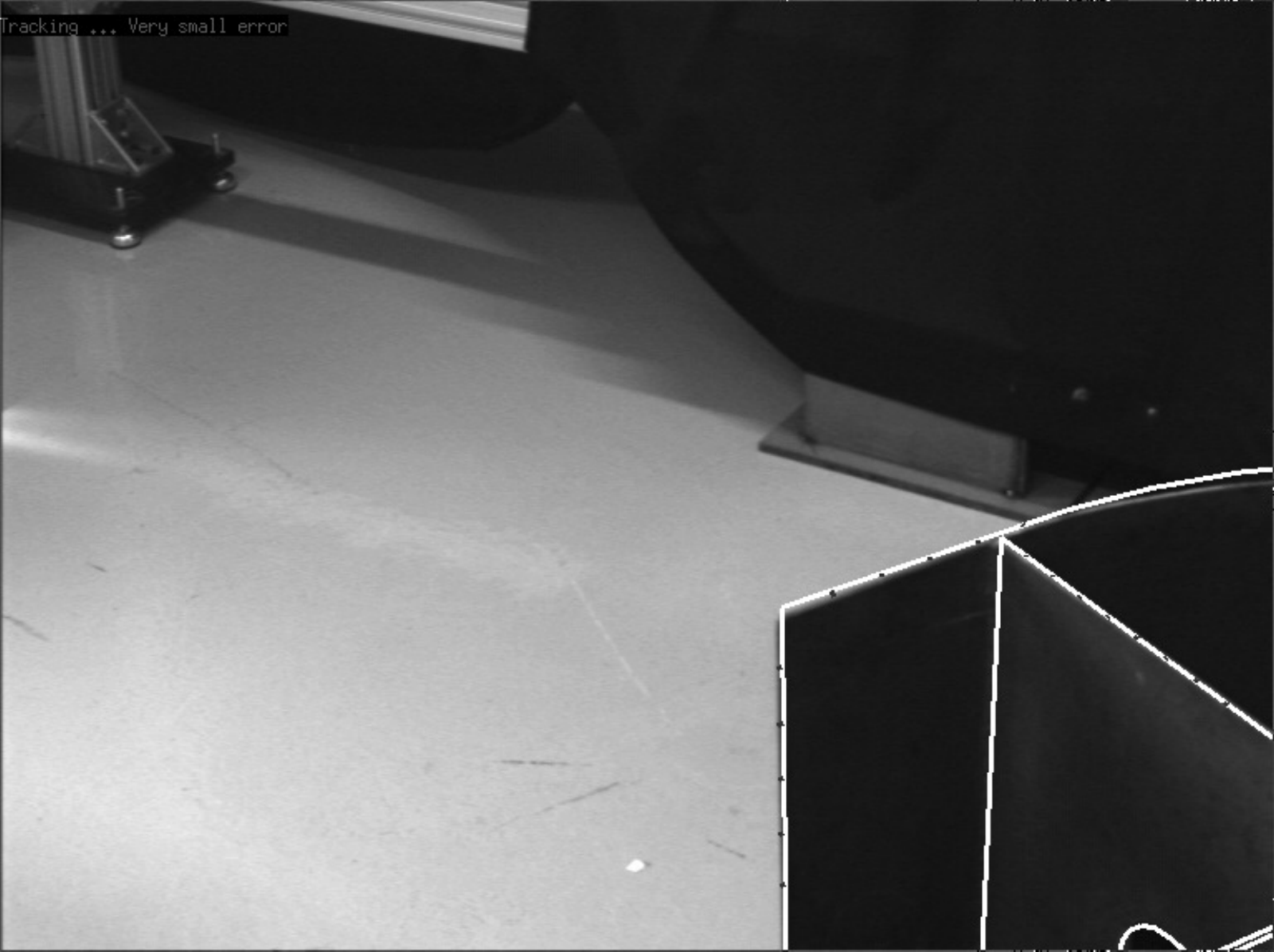}}&\framebox{\includegraphics{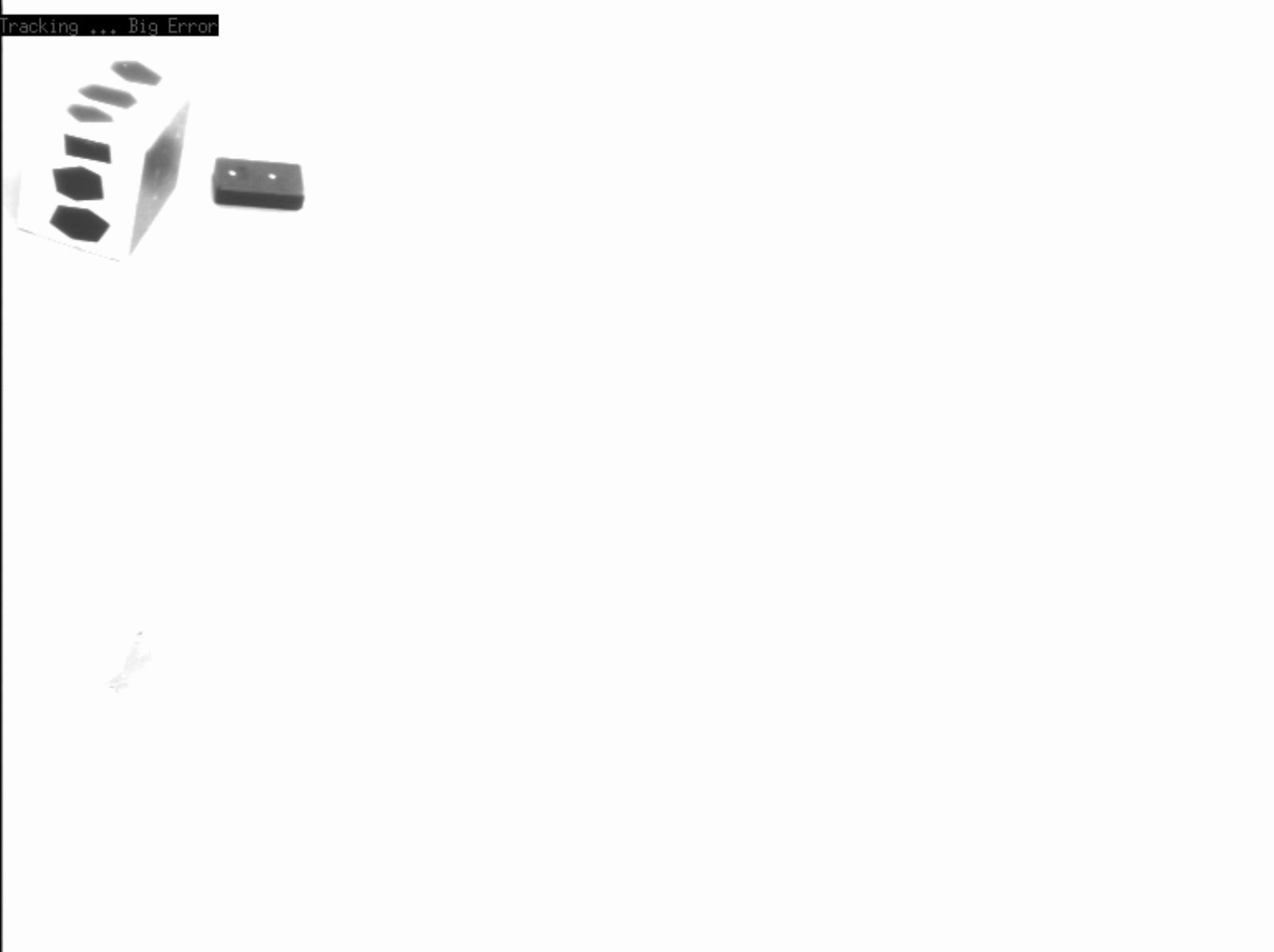}}\\
\end{tabular}
}}
\caption{Total occlusion in one image and partial occlusion in two images.}
\label{fig:tracking-occlusion1}
\end{figure}

\begin{figure}
\centerline{
\resizebox{\textwidth}{!}{\begin{tabular}{c@{\hspace{8mm}}c@{\hspace{8mm}}c}
\framebox{\includegraphics{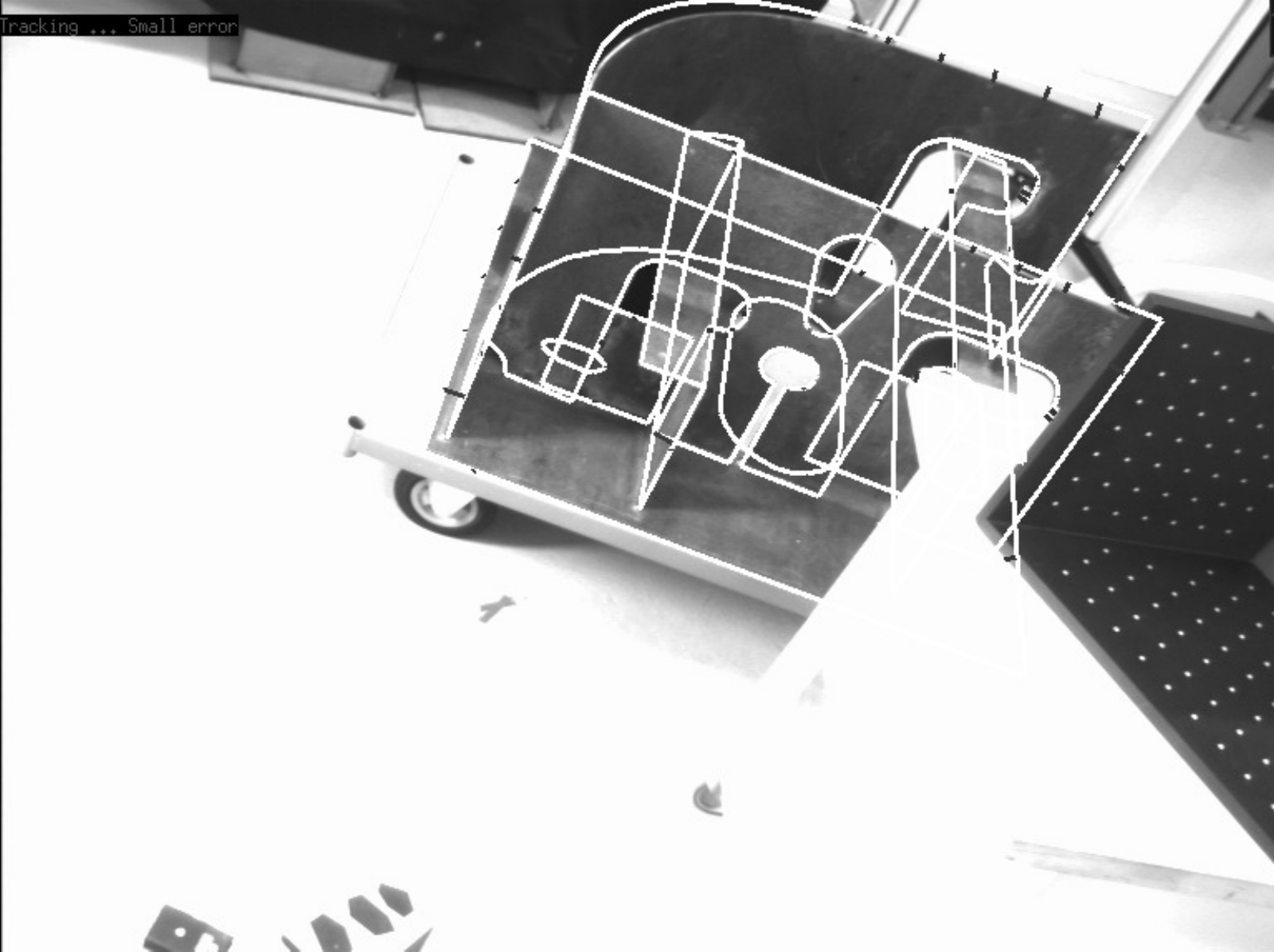}}&\framebox{\includegraphics{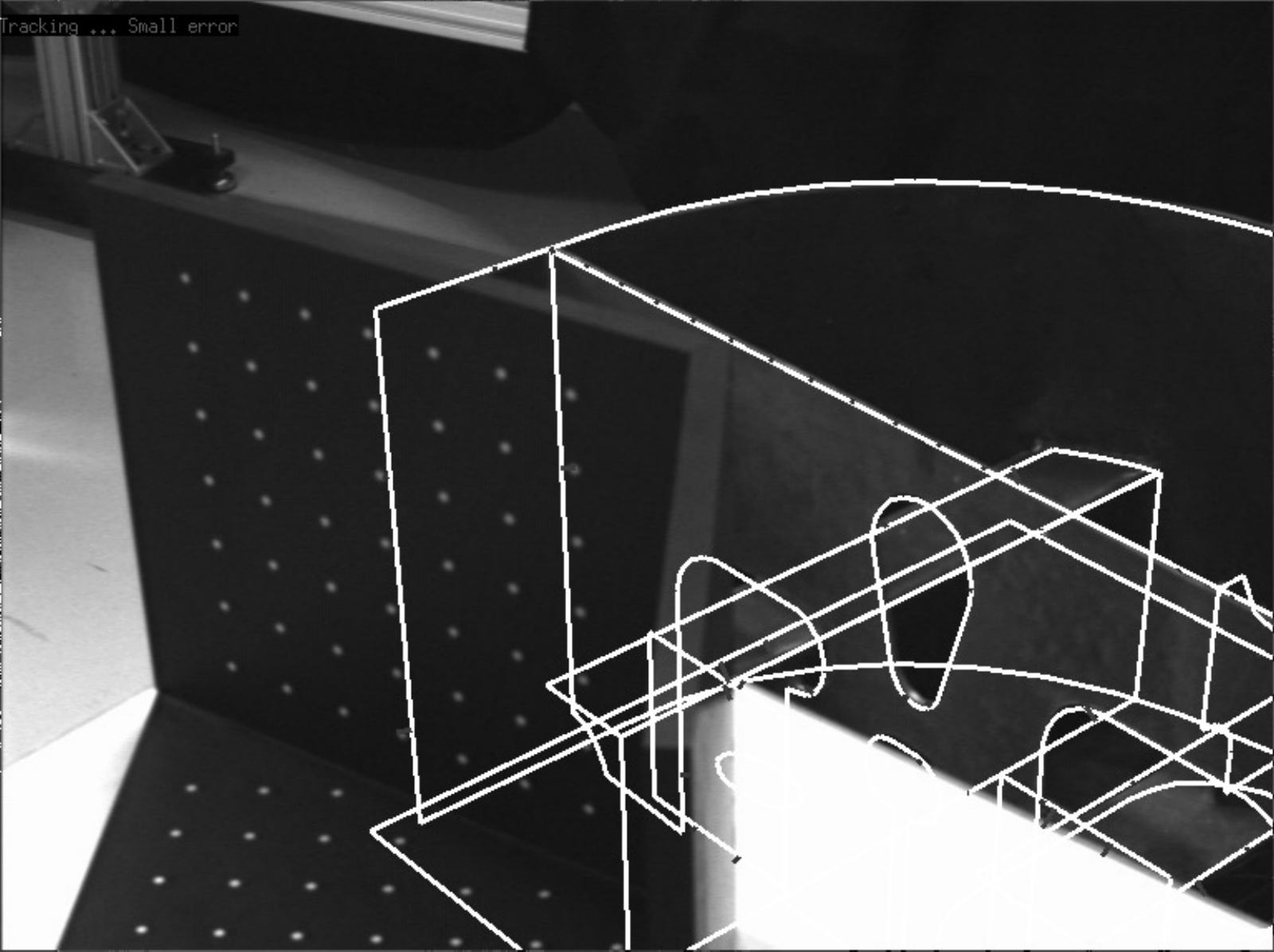}}&\framebox{\includegraphics{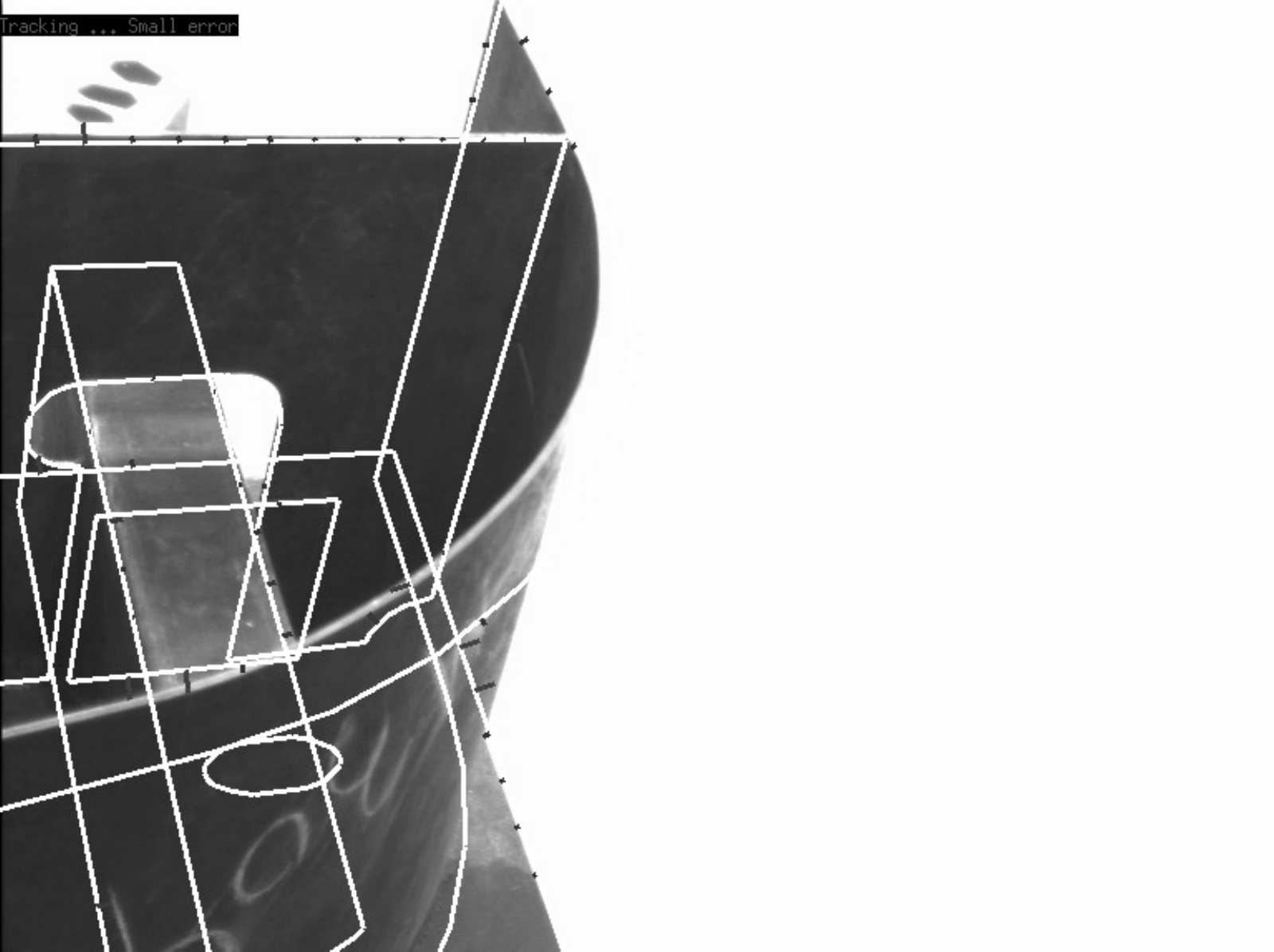}}\\
\end{tabular}
}}
\caption{Partial occlusion in three images.}
\label{fig:tracking-occlusion2}
\end{figure}

The next three figures show the result of tracking with three
cameras. Figure~\ref{fig:tracking-3cameras} shows a typical tracking
sequence where the object freely moves onto the ground. Notice the
excellent overlapping between model and object through the whole
sequence in spite of the fact that the object travels away from the
field of view of two cameras. Figure~\ref{fig:tracking-occlusion1}
shows partial occlusion in two images and total occlusion in one image
while Figure~\ref{fig:tracking-occlusion2} shows partial occlusion in
all three images. 

\section{Discussion}
In this paper we described a method for tracking rigid objects using
one or several cameras. The tracking process consists of aligning a
3-D model representation of an object with images by measuring
and minimizing the image error between predicted model contours and
image edge points. The tracker behaves like a visual servo loop where the
internal and external camera parameters are updated at each new image
acquisition.

We studied in detail the Jacobian matrix associated with this
minimization process in the presence of both point-to-point and
point-to-contour matches. We established the minimal number of matches
that are needed as well as the singular configurations leading to a
rank-defficient Jacobian matrix. We established a mathematical link
between the point-to-point and point-to-contour cases. Based on this
link we showed that the latter has the same kind of singularities than
the former. Moreover, we studied multiple camera configurations which
optimize the robustness of the method in the presence of single-camera
singularities, bad, noisy, or missing data.

Extensive experiments done with a complex ship part and with up to
three cameras validated the method. In particular we show that the
tracker may well be used as a camera calibration procedure.

One limitation of the method is that it only deals with edges. In
complex natural and industrial environments an edge-based method has
failures. Therefore we plan to incorporate texture information with
our model representation. When a model edge is predicted in the image,
nearby model texture is predicted as well and hence it is possible to
measure the discrepancy with image texture and tune the tracked
parameters by optimizing the alignment between model and image
textures. 

The fusion of several cues such as edge and texture information,
combined with the increasing computing power available today, is
likely to improve the robustness and realiability of visual trackers for many robotics
applications.

\printbibliography

\end{document}